\documentclass[sigconf, screen]{acmart}
\usepackage[ruled,linesnumbered]{algorithm2e}
\usepackage{subcaption}
\usepackage{booktabs}
\usepackage{multirow}
\AtBeginDocument{%
  }

\setcopyright{acmcopyright}
\copyrightyear{2018}
\acmYear{2018}
\acmDOI{XXXXXXX.XXXXXXX}

\acmConference[Conference unknown]{Make sure to enter the correct
  conference title from your right confirmation email}{unknown date}{unknown city}

\acmPrice{15.00}
\acmISBN{978-1-4503-XXXX-X/18/06}

\begin{document}

\title{Membership inference attack with relative decision boundary distance}

\author{JiaCheng Xu}
\email{2011263@tongji.edu.cn}
\affiliation{%
  \institution{Tongji University}
  \streetaddress{}
  \city{Shanghai}
  \state{}
  \country{China}
  \postcode{}
}
\author{ChengXiang Tan}
\authornotemark[1]
\email{jerrytan@tongji.edu.cn}
\affiliation{%
  \institution{Tongji University}
  \streetaddress{}
  \city{Shanghai}
  \state{}
  \country{China}
  \postcode{}
}

% \renewcommand{\shortauthors}{Anonymous Author, et al.}

%% delete the statement
\settopmatter{printacmref=false}
\renewcommand\footnotetextcopyrightpermission[1]{}

\begin{abstract}
Membership inference attack is one of the most popular privacy attacks in machine learning, which aims to predict whether a given sample was contained in the target model's training set.
Label-only membership inference attack is a variant that exploits sample robustness and attracts more attention since it assumes a practical scenario in which the adversary only has access to the predicted labels of the input samples.
However, since the decision boundary distance, which measures robustness, is strongly affected by the random initial image, the adversary may get opposite results even for the same input samples.

In this paper, we propose a new attack method, called muti-class adaptive membership inference attack in the label-only setting.
All decision boundary distances for all target classes have been traversed in the early attack iterations, and the subsequent attack iterations continue with the shortest decision boundary distance to obtain a stable and optimal decision boundary distance.
Instead of using a single boundary distance, the relative boundary distance between samples and neighboring points has also been employed as a new membership score to distinguish between member samples inside the training set and nonmember samples outside the training set.
Experiments show that previous label-only membership inference attacks using the untargeted HopSkipJump algorithm fail to achieve optimal decision bounds in more than half of the samples, whereas our multi-targeted HopSkipJump algorithm succeeds in almost all samples.
In addition, extensive experiments show that our multi-class adaptive MIA outperforms current label-only membership inference attacks in the CIFAR10, and CIFAR100 datasets, especially for the true positive rate at low false positive rates metric.
\end{abstract}


%   <ccs2012>
%   <concept>
%   <concept_id>10002978</concept_id>
%   <concept_desc>Security and privacy</concept_desc>
%   </concept>
%   <concept>
%   <concept_id>10010147.10010257</concept_id>
%   <concept_desc>Computing methodologies~Machine learning</concept_desc>
%   </concept>
%   </ccs2012>

% \ccsdesc[500]{Security and privacy}
% \ccsdesc[300]{Computing methodologies~Machine learning}

%     seats. Ichiro Suzuki preparing to bat.}

% \received{}
% \received[revised]{}
% \received[accepted]{}

\maketitle

\section{INTRODUCTION}
Machine learning algorithms have been trained on increasingly sensitive or private information such as medical history \cite{fernando2021deep,yu2021reinforcement}, political orientation \cite{cardaioli2020predicting,beltran2021male}, and criminal records \cite{wexler2019if,sagala2022comparative}.
However, recent research has shown that trained models can unexpectedly reveal the personal information \cite{song2020overlearning,mehnaz2022your,balle2022reconstructing}.
As one of the simplest forms of these information leaks, membership inference attacks (MIA) have attracted much attention.
For a given sample and a trained target model, the purpose of MIA is to infer whether it is a member sample involved in model training or a non-member sample outside the training set.

\begin{figure}[htp]
  \centering
  \begin{minipage}{0.9\linewidth}
    \centering
    \includegraphics[width=0.9\linewidth]{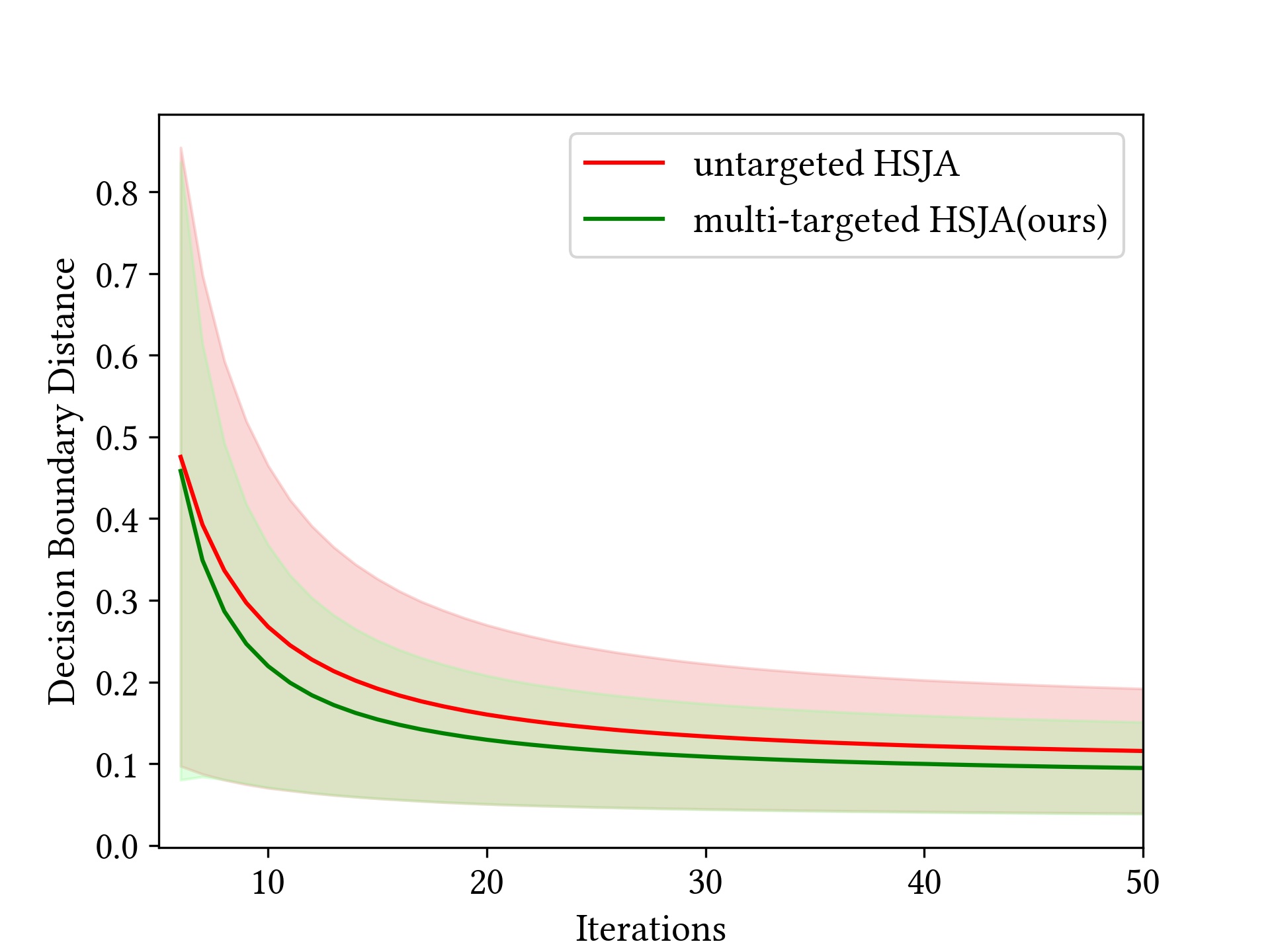}
    \subcaption{Decision boundary distances in different iterations.}
    \label{pt:comparison_with_old(a)}
  \end{minipage}
  \qquad
  \begin{minipage}{0.9\linewidth}
    \centering
    \includegraphics[width=0.9\linewidth]{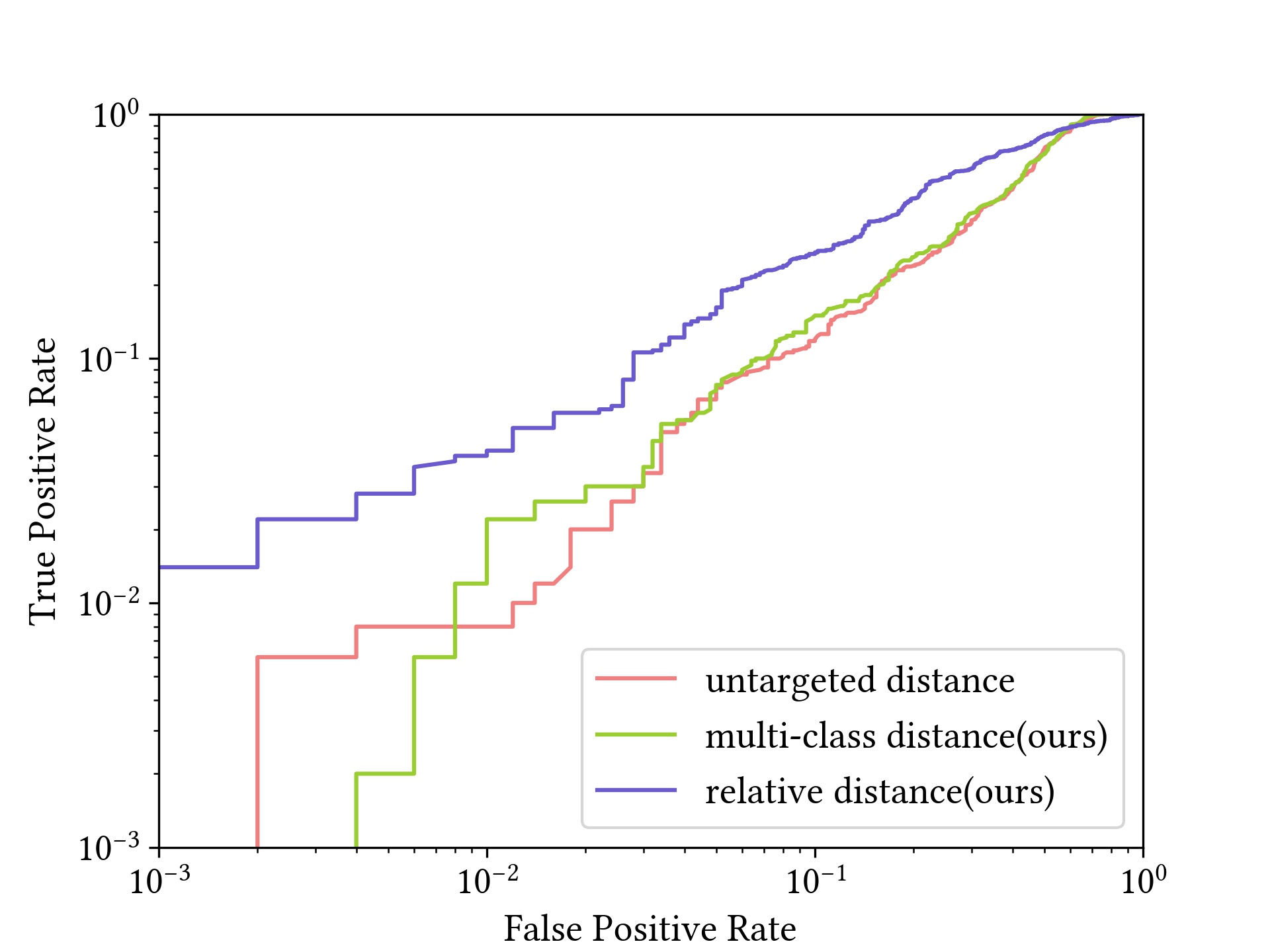}
    \subcaption{Full log-scale ROC curves for different attack performance.}
    \label{pt:comparison_with_old(b)}
  \end{minipage}
  \caption{(a)Decision boundary distances with two HSJA algorithms in different iterations; 
           (b)Full log-scale ROC curves of MIA with different membership score choices}
  \label{pt:comparison_with_old}
  \Description{A scatter.}
\end{figure}
Most MIAs can be categorized as score-based MIAs \cite{shokri2017membership,carlini2022membership,ye2022enhanced} because these attacks require the target model to output a confidence score for each sample as a precondition \cite{liu2022ml}.
The intuition behind these attacks is that a member sample is more likely to achieve a high confidence score both in target models or transfer models. 

In other words, confidence score is regarded as the membership score to distinguish members from nonmembers.
However, it seems impractical that the adversary can always obtain confidence scores, and some defensive strategies can also mislead MIA by using confidence values of adversarial samples instead, such as MemGuard\cite{jia2019memguard} and adversarial regularization\cite{nasr2018machine}.

These restrictions have motivated a more practical and strict MIA, also known as the label-only MIA \cite{choquette2021label,li2021membership,del2022leveraging}, in which the target model returns only the predictive labels of the samples (hard labels) and the adversary does not have any information about the internal computation of the model, such as confidence values.
The robustness of each sample has been taken as a proxy for the confidence value, and the intuition behind this is that member samples are more difficult to perturb than non-member samples\cite{choquette2021label,li2021membership}.
To evaluate sample robustness in the presence of hard labels only, these label-only attacks introduced the decision boundary distance as the new membership score, which is obtained by the untargeted HopSkipJump Attack (HSJA) algorithm\cite{chen2020hopskipjumpattack}.
HSJA is a variant of boundary attack \cite{brendel2017decision} that estimates the gradient direction of the decision boundary, descends along the gradient, and iterates the above process to gradually approach the minimum perturbed point of the decision boundary distance.
Unfortunately, although these attacks achieved good performance in several datasets, they failed to obtain stable and minimum decision boundary distances, and their evaluation methods do not reflect the privacy risks of individual samples.

In this paper, we re-examine the problem of current label-only MIAs and propose a novel one with a more reliable evaluation methodology.
We first argue that over half of the decision boundary distance generated by untargeted HSJA may not be reliable since the initial perturbed image is generated randomly.
Vo et al. \cite{vo2021ramboattack} has proved that various initial images could make a vital difference in minimum perturbed distance under the same source image.
Hence, we propose a multi-targeted HSJA instead of untargeted HSJA to generate the stable and minimum decision boundary distance for each sample.
As illustrated in Figure \ref{pt:comparison_with_old(a)}, using the same evaluation dataset, our method produces smaller and more stable decision boundary distances for any number of iterations.

Besides the decision boundary distance, our second argument is that the performance of the traditional label-only MIA attackers was overestimated on the commonly used balanced set, especially for the overfitted target model.
We observed that these attacks, beyond their claimed attack strategies, also performed another baseline attack\cite{yeom2018privacy} on misclassified samples and achieved much higher attack performances than themselves.
In other words, their results evaluated for a balanced set consisting of the same number of members and nonmembers were essentially the results of hybrid attacks which could not reflect the performance of their attacks own. 
Even worse, their attack performance was further overestimated as the model overfitting increases and more misclassified samples appeared in the balanced set.
Therefore, we propose the cbalanced set, which is a balanced set with all samples correctly classified, as the new evaluation set, and take true positive rate (TPR) at low false positive rate (FPR) and full log-scale ROC curve as the new evaluation metrics.

When evaluated this way, we find most prior label-only MIAs fail to distinguish members from nonmembers due to their close decision boundary distance distributions.
The main challenge for these attacks issue is to distinguish between hard nonmembers (for which boundary distance is high) and easy members (for which boundary distance is low).
Similar challenge has already been solved in the score-based setting with sample calibration\cite{watsonimportance,carlini2022membership,rezaei2022efficient}, however their solutions require confidence scores to train a number of shadow models, which is hard to be implemented in label-only settings.
Therefore, we propose a novel relative decision boundary distance as a membership score that implements an adaptive decision boundary distance threshold for each sample.
This relative decision boundary distance calculates the average difference between the sample and neighboring point decision boundary distances for sample calibration\cite{watsonimportance}, without any shadow models\cite{carlini2022membership} or sample subpopulations\cite{rezaei2022efficient}.
Compared to the single boundary distance, as shown in the Figure \ref{pt:comparison_with_old(b)}, the relative boundary distance as the membership score achieves a higher TPR at a lower FPR, demonstrating that our method is able to detect more member samples at low error rates.
Abstractly, our contribution can be summarized as follows:
\begin{itemize}
  \item We first point out that the decision boundary distances obtained by conventional untargeted HSJA are not stable due to the difference in their random initial image classes, 
        and propose a multi-targeted HSJA to obtain stable and minimum decision boundary distances.
  \item We explore the shortcoming of the balanced set in traditional evaluation datasets and take a more restricted cbalanced set as
        the evaluation set and low false-positive rates as the evaluation metric.
  \item We analyze the failure of prior label-only MIAs in new evaluation methodology
        and propose a novel label attack with relative decision boundary distances as membership score.
\end{itemize}

\section{PRELIMINARY}
\subsection{Machine Learning}
For supervised machine learning classification tasks, let $M$ denotes the learned neural network, which maps samples from a dataset $\mathbf{X}$ to classes in a label set $\mathbf{Y}$.
Given a sample $x$, $M$ outputs a vector $y=M(x)=(y_{1},y_{2},...,y_{n})$ and $y_{i}$ represents the prediction probability of i-th class.
Based on this vector, the confidence value $y^{c}$ and the predicted class $y^{l}$ can be achieved in the following:
\begin{equation}
  \label{eq1}
  y^{c} = softmax(y) = (y_{1}^{c}, y_{2}^{c},...,y_{n}^{c}),y_{i}^{c}=\frac{e^{y_{i} } }{ {\textstyle \sum_{j}^{n}e^{y_{j} } } }
\end{equation}
\begin{equation}
  \label{eq2}
  y^{l} = \mathop{\arg\max}\limits_{i}y_{i}^{c}
\end{equation}
\subsection{Membership Inference Attack}
The objective of MIA is to be capable of distinguishing member samples in the training dataset from nonmember samples outside the training dataset.
Since the first MIA has been proposed by Shokri et al. \cite{shokri2017membership}, a number of attack and defense mechanisms have been developed until now.
The essence of MIA is not concerned with the leakage of the specific value of a sample, but with the leakage of the membership of a sample, i.e., whether a sample increases the probability of being identified as a member by participating in the model training, similar to differential privacy\cite{dwork2006differential}.
Hence, MIA can also be seen as one of the smallest and simplest forms of privacy leakage, which is defined as follows:
\paragraph{Definition of MIA}
Given a sample $x$, a trained ML model $M$, a training dataset $D_{train}$ and some auxiliary information denoted as $I$,
we follow a common definition of membership inference attack $A$ just as follows\cite{liu2022membership}:
\begin{equation}
  A:x,M,I \to \left \{ 0,1 \right \}
\end{equation}
where 1 means $x \in D_{train}$ and 0 means $x \notin D_{train}$.
\paragraph{Evaluation of MIA}
Since the output of MIA is binary, the evaluation of MIA can be summarized as a classical inference game\cite{yeom2018privacy}, and then the attack performance has been measured with balanced set and average-case metrics in most studies.
However, Carlini et al. \cite{carlini2022membership} has argued that average-cases metrics such as area under curve(AUC) are nonsense at a high FPR and suggested TPR at low FPR instead.
Similarly, Jayaraman et al. \cite{jayaraman2021revisiting} has noticed that balanced set consisting of the member and nonmembers in the same size is impractical and suggested imbalanced dataset instead.
In this paper, we both take TPR at low FPR and AUC as our evaluation metrics for the integrity, and we take an advanced balanced set for our evaluation dataset.
The details of advanced balanced dataset are in Section \ref{subsection:METHOD-1}.
\paragraph{Label-only MIA}
To better evaluate the actual performance of MIA, apart from the evaluation of the attack result, it is also necessary to consider the attack precondition, i.e., the adversary's access to the target model.
In most cases, the confidence values $y^{c}$ was considered a requisite for the adversary, which is also called score-based MIA.
However, confidence values are not as easy to obtain and defenders can easily mislead score-based attacks by modifying the confidence values.
Therefore, several researches proposed the label-only MIA\cite{choquette2021label,li2021membership}, whose adversary only have the query access to the model and only require the hard predicted label $y^{l}$.
These label-only MIAs generate a perturbed point $x_{adv}$ for each data point $x$ and consider the minimum norm distance $l_{p}$ between $x_{adv}$ and $x$ as a proxy for confidence score.
Similar to the intuition that members are more likely to achieve a high confidence score than nonmembers behind the score-based MIA, the intuition behind these label-only MIA is that members are harder to be perturbed than nonmembers.
The details of the intuition can be seen in Section \ref{subsection:METHOD-2}

\section{METHOD}
\label{section:METHOD}
In this section, we introduce multi-class adaptive MIA, a novel label-only MIA with multiple initial images for each target class and an adaptive membership score for each sample.
We start by outlining the threat model and estimation mechanism.
Then we describe the rationale behind our attack and explain why it is effective.
Finally, we go on to the specifics of our attack pipeline and show how it is organized.
\subsection{Threat Model And Estimation Mechanism}
\label{subsection:METHOD-1}
In this paper, we concentrate on a practical, black-box MIA scenario in which the target model solely outputs the hard label results without providing any confidence score.
Furthermore, we suppose that the query cost places a limit on the number of requests an adversary can send. 
The aforementioned threat model complies with the requirements for the majority of label-only MIAs\cite{choquette2021label,li2021membership}, while our evaluation procedures differ from the traditional methodology.

\paragraph{Evaluation Dataset}
The majority of label-only MIAs employed a balanced dataset, consisting of an equal amount of data picked at random from the training and test sets, as the evaluation dataset to assess how frequently an attack correctly predicts membership.
However, we contend that this balanced dataset overestimated the underlying membership leakage of prior label-only attacks.
The main cause of this overestimation is that these attacks preprocess misclassified samples by treating them directly as nonmembers with zero decision boundary distance, which is also called the baseline attack\cite{yeom2018privacy}.
As a result, attacks on incorrectly classified samples generally work much better than those on correctly classified samples.
Even worse, when the model is overfitted, the proportion of incorrectly classified samples in the balanced set rises, and the effectiveness of the attack as a whole depends more on the success of the baseline attack.
Therefore, we propose a new c-balance dataset that retains the characteristics of the original balance dataset while requiring the same prediction labels for each sample point as the ground truth.
In a cbalanced set, the baseline attack's accuracy is only 50\%, the same as the accuracy of a random prediction, hence maximizing the recovery of the attack's own strategy's performance.

\paragraph{Evaluation Metric}
Most label-only MIAs measured the effectiveness of their attacks using average-case metrics like AUC and accuracy.
However, Rezaei et al. \cite{rezaei2021difficulty} pointed out that these metrics paled in comparison to the significance of the false positive rate, which refers to nonmembers who are misclassified as members.
Carlini et al. \cite{carlini2022membership} also mentioned that it was common practice in many computer security domains to build solutions with low false positive rates.
A simple illustration is that when both attacks have an AUC of 0.7, the attacker will typically opt for the attack with the higher true positive rate (TPR) and lower false positive rate (FPR), as the FPR above 0.5 has no real impact on the privacy risk.
Therefore, we used the TPR at low FPR and the ROC curve on a logarithmic scale as new evaluation metrics and retained the AUC metric only for completeness of comparison.

\subsection{Attack Intuition}
\label{subsection:METHOD-2}
The objective of MIA is to find a powerful and explainable membership score function to distinguish members and nonmembers.
For a given sample $x$, a trained target model $M$, a predicted probability $y^{c}$, and a ground truth $y^{t}$, a general membership score function $s_{mem}$ can be represented in the following:
\begin{equation}
  s_{mem}\left(x, y^{c}, y^{t}\right) = l\left(x, y^{c}, y^{t}\right), y^{c}=M\left(x\right)
\end{equation}
where $l$ denoted the model loss function.
The intuition behind MIAs is that since the model is trained to minimize the loss function, the loss of member samples is on average less than that of non-members.
When it comes to label-only MIAs, most of them has assumed that an adversary has no access to $l$ and $y^{c}$, and $s_{mem}$ is represented in the following instead:
\begin{equation}
  \begin{aligned}
     & s_{mem}\left(x, y^{t}\right) = \min_{x_{p} \in P } d\left ( x, x_{p} \right )                       \\
     & P=\left \{ x_{p} \mid \mathop{\arg\max}\limits_{}M\left ( x_{p} \right )= y^{t} \right \} \nonumber
  \end{aligned}
\end{equation}
where $d$ denotes the minimum norm distance between $x$ and $x_{p}$, $x_{p}$ samples from perturbed samples of $x$, which has the same hard label with $x$.
\begin{figure}[htp]
  \centering
  \includegraphics[width=\linewidth]{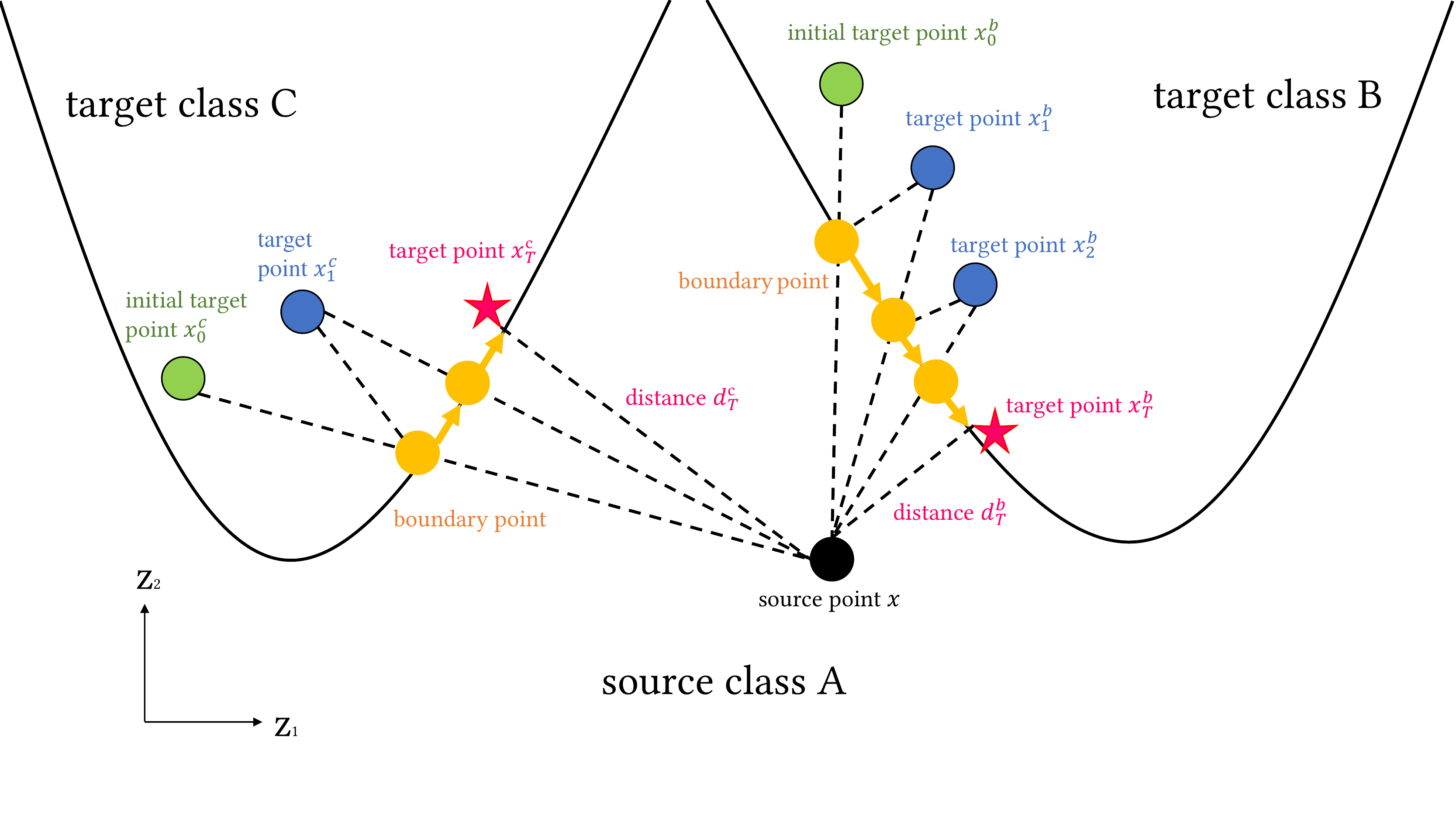}
  \caption{2D($z_{1}$ and $z_{2}$) Input Space Example. An illustration of the untargeted HSJA algorithm's decision boundary distances at the initial target points of various target classes.}
  \label{pt:2D example}
  \Description{A scatter.}
\end{figure}

The intuition behind label-only MIAs is that members are more likely to have longer decision boundary distances on average than nonmembers since models are trained to accurately classify members and to be away from the decision boundary.
However, motivated by the difficulty calibration \cite{watsonimportance} in score-based settings, we argue that a considerable part of the decision boundary distance of member samples is shorter than that of some non-member samples due to the diversity of samples, especially in some models with good generalization.

As a result, our new attack intuition is that the relative decision boundary distances of members are generally longer than those of nonmembers.
Assuming that the neighboring points can be identified by small translations from the original sample point, the relative decision boundary distance can be defined as the expectation of the difference in decision boundary distance between an original sample point and its neighboring points.
Our attack approach also makes a basic assumption that the decision boundary distance expectation of multiple neighboring points can be equivalent to the decision boundary distance when that sample point is a nonmember.
According to this assumption, the original sample, if it is a nonmember, should have a shorter relative decision distance than a member since it belongs to the same nonmember distribution as the neighboring points.
Our strategy can effectively address the issue of sample diversity since a nonmember with a large decision boundary distance is also likely to have several neighboring points with large decision boundary distances, leading to a short relative decision boundary distance.
\subsection{Attack Method}
\begin{figure*}[htp]
  \centering
  \includegraphics[width=\linewidth]{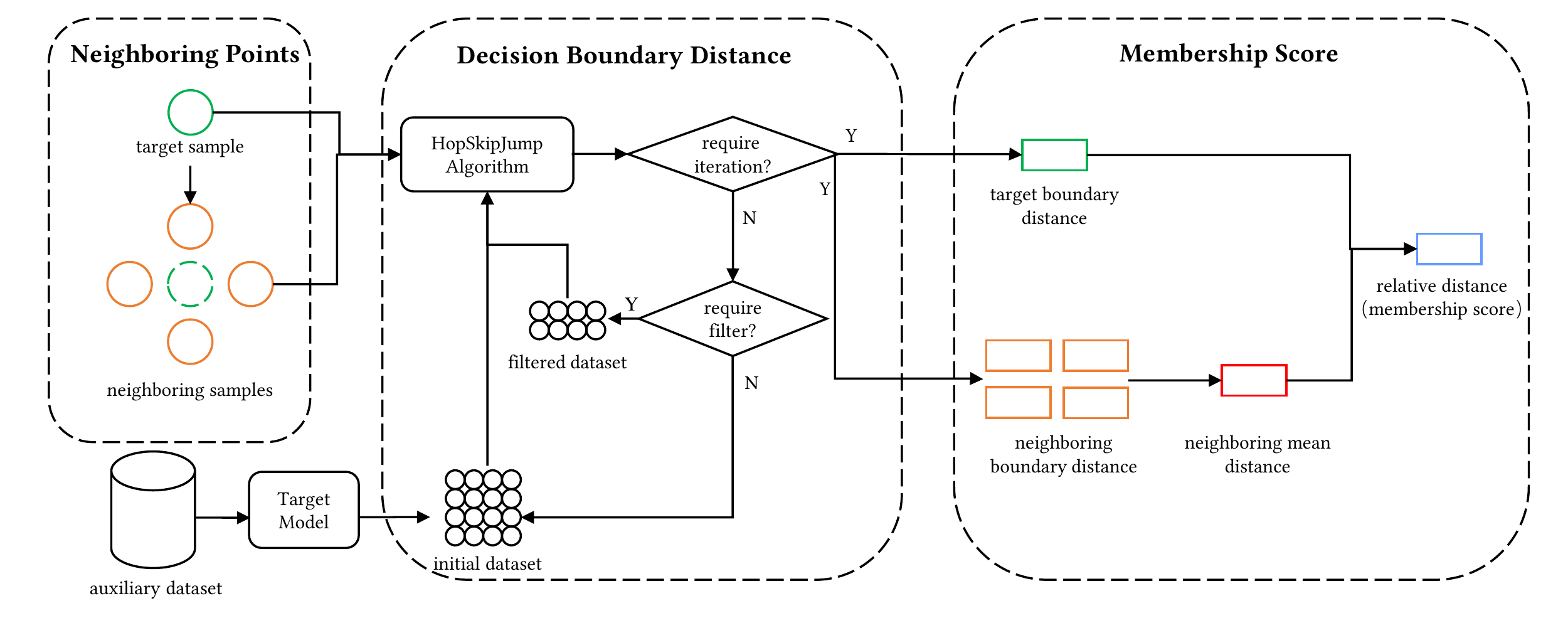}
  \caption{Our attack framework.}
  \label{pt:attack framework}
  \Description{Attack framework.}
\end{figure*}

In this paper, we propose a novel label-only membership inference attack, called multi-class adaptive MIA.
The detailed pipeline of our attack, which includes neighboring points, decision boundary, and membership score, is shown in figure \ref{pt:attack framework}.
In label-only circumstances, neighboring points searching, and adaptive membership score are first introduced.
The HSJA algorithm for determining decision boundary distance is similar to the conventional label-only MIA, with the exception that stable and minimum decision boundary is obtained by accounting for numerous initial classes.
\paragraph{Neighboring points}
As stated in our attack intuition, in order to imitate the performance of the original points in the situation of nonmembers, neighboring points must be kept close to the original points. 
Therefore, we slightly translate each sample image to locate neighboring points. 
\begin{algorithm}
  \SetAlgoLined
  \KwIn{Adversarial algorithm $HopSkipJump$, a target model $M$, a candidate sample $(x,y)$, iteration $T$}
  \KwOut{minimum perturbed sample $x_{p}$}
  initial a perturbed $x_{0}$ with adding random noise to $x$, make $M(x_{0}) \ne y$\;
  \For{$t$ in $1,2,...,T$}{
    $x_{t} = HopSkipJump\left(x,x_{t-1},M,t\right)$}
  return $x_{T}$\;
  \caption{untargeted HSJA}
  \label{alg:untargeted hsja}
\end{algorithm}
Empirically, up and down and left and right are chosen as the translation directions, and a distance of one is used.
\paragraph{Decision boundary}
After searching for all the neighboring points, we need to obtain the decision boundary distance for each sample point or neighboring point to measure their robustness.
The majority of label-only MIAs use the untargeted HSJA algorithm to generate a minimum perturbed point on the decision boundary, and take the norm distance between the original point and the minimum perturbation point as the decision boundary distance.
The untargeted HSJA algorithm, as depicted in the Figure \ref{pt:2D example} and Algorithm \ref{alg:untargeted hsja}, first generates the initial point of the target class randomly, which is distinct from the source class, then obtains the boundary point via binary search, estimates the gradient, and progresses geometrically to obtain the new target point. 
It then repeats the last two operations until it reaches the maximum number of iterations, at which point it obtains the minimum perturbed point.

% HSJA is a popular algorithm that divides into two types: targeted attacks with a fixed initial image and untargeted attacks with random images. 
% It works by estimating the decision boundary gradient and descending along the gradient to achieve the minimum perturbed point on the boundary.
However, Vo et al. \cite{vo2021ramboattack} has recently shown that the minimum perturbed point is significantly influenced by the initial image selection, which prompts us to rethink the stability of the obtained boundary distances.
We repeat untargeted HSJA ten times for the same samples and find that only around half of them manage to get close to the minimum perturbed point. 
After thorough testing and analysis, we discovered that the minimum perturbed point and the initial target class had a significant relationship, just as shown in Figure \ref{pt:2D example} and more details are given in Section \ref{subsec: decision boundary}.

% Since almost all HSJA implemented in label-only MIAs were untargeted,  these randomly generated initial images may prevent the adversary from getting the true decision boundary distance.
% We observed that stable and accurate decision boundary distances were not available for more than half of the tested samples.
% We take a further step in these changed distances and find a strong relation between the changed distances and initial image labels.
% Therefore, we propose a hypothesis that the minimum perturbed point that HSJA can reach is strongly correlated with the label of the initial image.

As a result, traversing the initial images of all labels is required to determine the minimum perturbation points.
Unfortunately, since the initial images with the specified label are challenging to generate randomly, it may be difficult to accomplish such traversal under untargeted HSJA.
Therefore, we perform targeted HSJA under the fixed initial image of each label independently, which is denoted as all-targeted HSJA and take the shortest decision boundary distance as the final decision boundary distance.
Although all-targeted HSJA can achieve stable decision boundary distances, it is hard to apply to multi-class tasks since it requires n-1 times as many queries as untargeted HSJA (n denotes the total number of classes).

In response to such a dilemma, we propose a multi-targeted HSJA to achieve the minimum perturbed points with a limited number of queries.
Our attack algorithm has been shown in Algorithm \ref{alg:multi-targeted hsja}, which includes a filter in the loop to eliminate a subset of the target classes whose decision boundary distance drops slowly along the gradient.
This filtering is repeated for each $T_{f}$ cycle until only one element in $D_{ini}$ remains as the final target class.
The assumption behind our algorithm is that the target class that eventually achieves the shortest decision boundary distance is often significantly shorter than the decision boundary distance of other classes in the first few rounds.
Since the gradient descent distance decreases with the number of rounds in the HSJA algorithm, it is difficult to change the ordering of the decision boundary distances for various classes in later rounds.

\begin{algorithm}
  \SetAlgoLined
  \KwIn{Adversarial algorithm $HopSkipJump$, a target model $M$,
    a candidate sample $(x,y)$, an auxiliary dataset $D_{aux}$,
    total iteration $T$, filter iteration $T_{f}$, filter rate $r$}
  \KwOut{minimum perturbed sample $x_{p}$}
  Sampling one correctly classified sample from each class in $D_{auc}$ to form a candidate initial dataset $D_{ini}$.\;
  \For{$t$ in $1,2,...,T$}{
    \For{$x^{p}$ in $D_{ini}$}{
      $x^{p}_{t} = HopSkipJump\left(x,x^{p}_{t-1},M,t\right)$\;
      $d^{p}_{t} = l_{2}\left(x,x^{p}_{t}\right)$\;
      }
    $D_{ini}=[x^{0}_{t},x^{1}_{t},\dots,x^{|D_{ini}|}_{t}]$\;
    \If{$t$ mod $T_{f}$ = 0}{
      sort $D_{ini}$ with $d^{p}_{t}$ in ascending order\;
      $D_{ini} = D_{ini}\left[0:r\right]$\;}
    }
  return $x^{0}_{T}$\;
  \caption{multi-targeted HSJA}
  \label{alg:multi-targeted hsja}
\end{algorithm}

\paragraph{Membership score}
After obtaining the decision boundary distances of sample points and neighboring points, we finally select the appropriate membership scores to distinguish members from nonmembers.
Traditional label-only MIAs directly take decision boundary distances as membership scores, use a fixed threshold to distinguish members from nonmembers, and perform poorly on correctly classified samples.
Aside from the previously noted instability of the decision boundary distance, the empirically fixed threshold also contributes to the failure of traditional methods.

The key factor causing these failures is that the average decision boundary distance of members is not considerably longer than that of nonmembers, especially when both are correctly classified.
As a result, a single fixed threshold cannot distinguish between members who have short decision boundary distances and nonmembers who have long decision boundary distances.

Similar problems have emerged and been solved in score-based settings\cite{carlini2022membership,ye2022enhanced}, however, their solutions are difficult to transfer straight to label-only settings because of the inability to train shadow models.
As a result, we present a new label-only membership inference attack in which the membership score is the relative decision boundary distance between a sample and its neighboring points.
The membership score in our attack is adjusted as follows:
\begin{equation}
  s_{ours}\left(x, y\right) =  s_{mem}\left(x, y\right) - \mathbb{E}_{x_{n}}s_{mem}\left(x_{n}, y_{n}\right)
\end{equation}
Due to the tiny translation distance, the expectation of the decision boundary distance of neighboring points can be approximated as the decision boundary distance when the original sample is not in the training set, as shown below:
\begin{equation}
  s_{ours}\left(x, y^{t}\right) \approx s_{mem}\left(x, y^{t}\right) - \mathbb{E}_{x \notin D_{train}}s_{mem}\left(x, y^{t}\right)
\end{equation}
The motivation is that the candidate sample's neighboring points must be nonmembers, and if the candidate sample is a nonmember, they are in the same distribution with closer decision boundary distances.
Essentially, the relative decision boundary distance denotes the change in decision boundary distance caused by a sample's absence from training. 
The likelihood that the sample is a nonmember increases with decreasing relative decision boundary distance.

\section{EVALUATION}
In this section, we conduct experiments on representative datasets with corresponding models to evaluate our multi-class adaptive MIA and compare it with other label-only MIAs.
The comparison is restricted to label-only MIAs since the amount of information retrieved per query in the label-only case is significantly less than that in the confidence value scenario.
\subsection{Experimental Setup}
\paragraph{Dataset} Our experiments focus on the following two datasets.
\begin{itemize}
  \item CIFAR-10\cite{krizhevsky2009learning}. The CIFAR-10 is a benchmark dataset for classification tasks and is also widely used in MIA evaluation.
        It consists of 60000 32 $\times$ 32 color images in 10 classes, with 6000 images per class.
  \item CIFAR-100\cite{krizhevsky2009learning}. This dataset is similar to CIFAR-10, except that it has 100 classes and each class contains 600 images.
\end{itemize}
\paragraph{Model} For the CIFAR-10 dataset, we construct the same model architecture as Li et al. \cite{li2021membership}, using 4 convolutional layers, 4 pooling layers, and 2 hidden layers with 256 units each as the final layer.
For the CIFAR-100 dataset, we build a model similar to ResNet-18, and due to the low image resolution, we changed the size of the first convolution kernel from 7 to 3 by referring to He et al\cite{he2016deep}.
Unless otherwise stated, a 5:1 ratio of training to test sets has been used to train all target models in this paper.
With a batch size of 128 and a learning rate of 0.001, the Adam algorithm is used to train both target models in 200 training epochs.
\paragraph{Metrics} We use the following evaluation metrics.
\begin{itemize}
  \item Log-scale ROC Curve. Receiver operating characteristic curves (ROC) are widely used to compare the true positive rate (TPR)
        and true positive rate (FPR) of attacks at all possible decision thresholds.
  \item TPR at low FPR. It records the attack performance at a single choice of low FPR (e.g., 0.1\%FPR) and facilitates a quick comparison of different attack configurations\cite{carlini2022membership}.
  \item AUC. It is an average-case metric widely used in a number of MIAs to measure the overall performance of classifiers for binary classification tasks.
        Although it proved to be unsuitable for assessing MIA in recent studies, we added it for completeness.
\end{itemize}
\paragraph{Baseline}
In terms of the decision boundary distance, we mainly compare our all-targeted HSJA and multi-targeted HSJA with the untargeted HSJA as a baseline.
Chen et al. \cite{chen2020hopskipjumpattack} first proposed the untargeted HSJA and targeted HSJA to find the minimum perturbed point by estimating the gradient descent.
In terms of the attack performance, we compare our multi-class adaptive MIA with two representative label-only MIA \cite{choquette2021label,li2021membership} as baseline.
Although Li et al. \cite{li2021membership} additionally used Query-efficient boundary-based attack(QEBA) \cite{li2020qeba} to generate minimum perturbed points, QEBA just reduced the query cost of the gradient estimation therefore these two representative attacks could be summed up as one untargeted MIA.
The performance of the conventional untargeted MIA on our cbalanced set is compared to that on the balanced set as a baseline for the evaluation set.

\subsection{Evaluation Dataset Comparison}
\label{subsec:evaluation set}
In this section, we compare the attack performance of untargeted MIA on the balanced set and the cbalanced set.

To assure the fairness of the comparison, we randomly picked 1000 samples from the same distributed dataset to generate a balanced set and a cbalanced set, respectively. 
Additionally, we conducted the comparison above on the target model with various levels of overfitting in order to account for the effect of overfitting on the MIA performance.
The degrees of model overfitting in different training size are illustrated in the train-test gap column in Table \ref{tab:overfitting with accuracy}.

\begin{figure*}[htp]
  \centering
  \begin{minipage}{0.32\linewidth}
    \centering
    \includegraphics[width=0.9\linewidth]{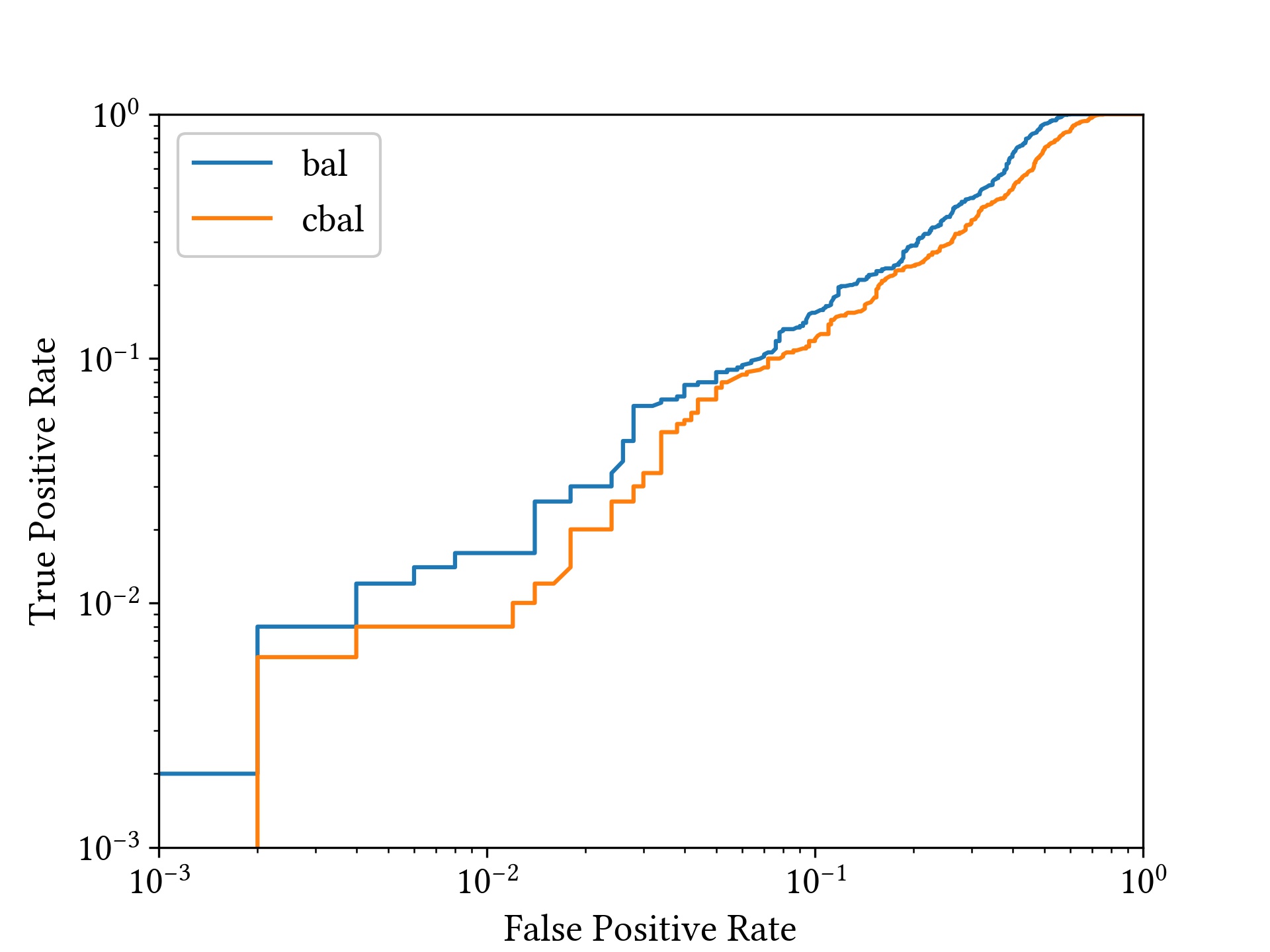}
		\caption*{(a)CIFAR10-M1} %?????subcaption????????????????????????????????????
		\label{pt:overfitting ROC(a)}%??????
  \end{minipage}
  \begin{minipage}{0.32\linewidth}
    \centering
    \includegraphics[width=0.9\linewidth]{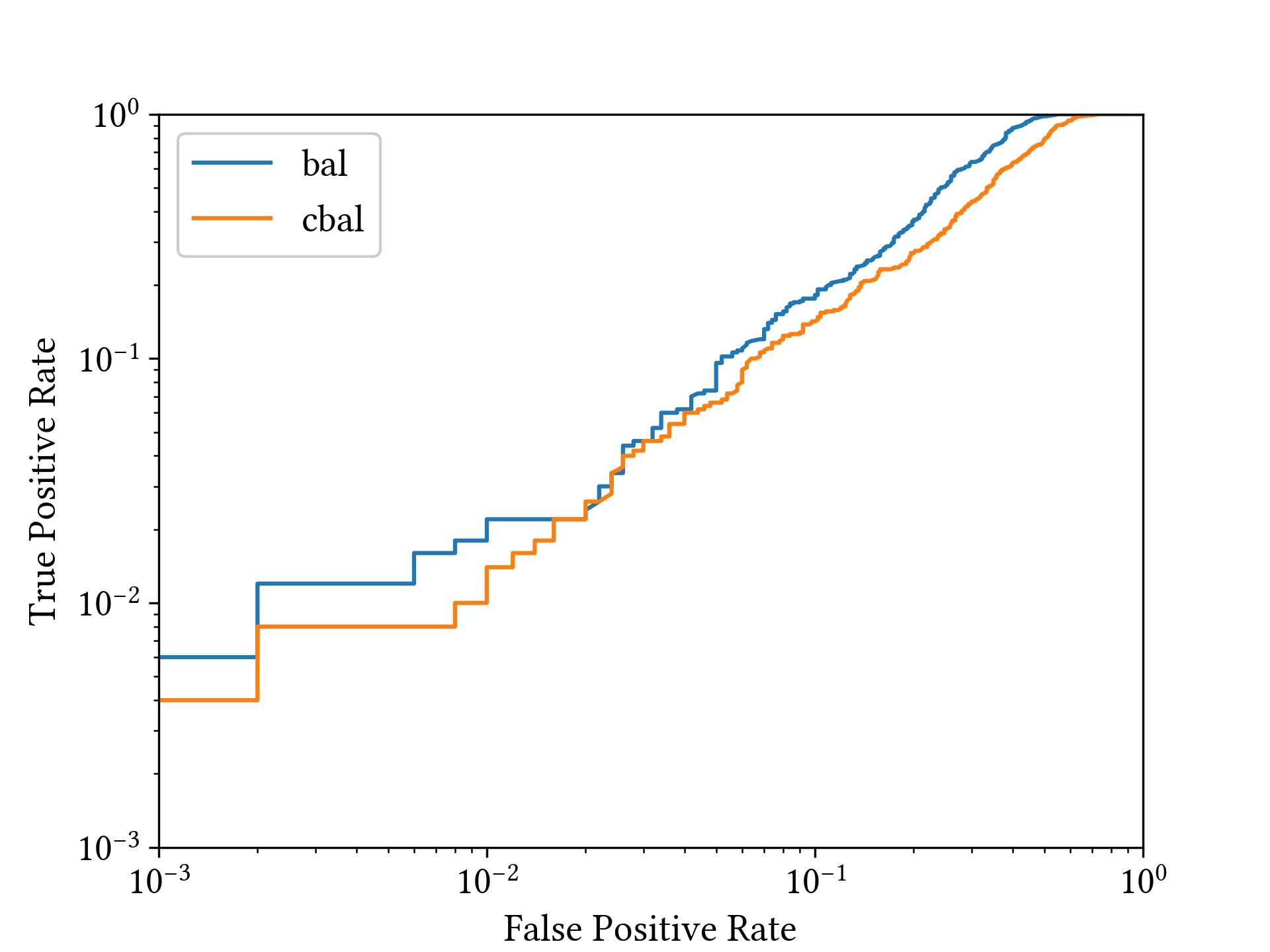}
		\caption*{(b)CIFAR10-M2} %?????subcaption????????????????????????????????????
		\label{pt:overfitting ROC(b)}%??????
  \end{minipage}
  \begin{minipage}{0.32\linewidth}
    \centering
    \includegraphics[width=0.9\linewidth]{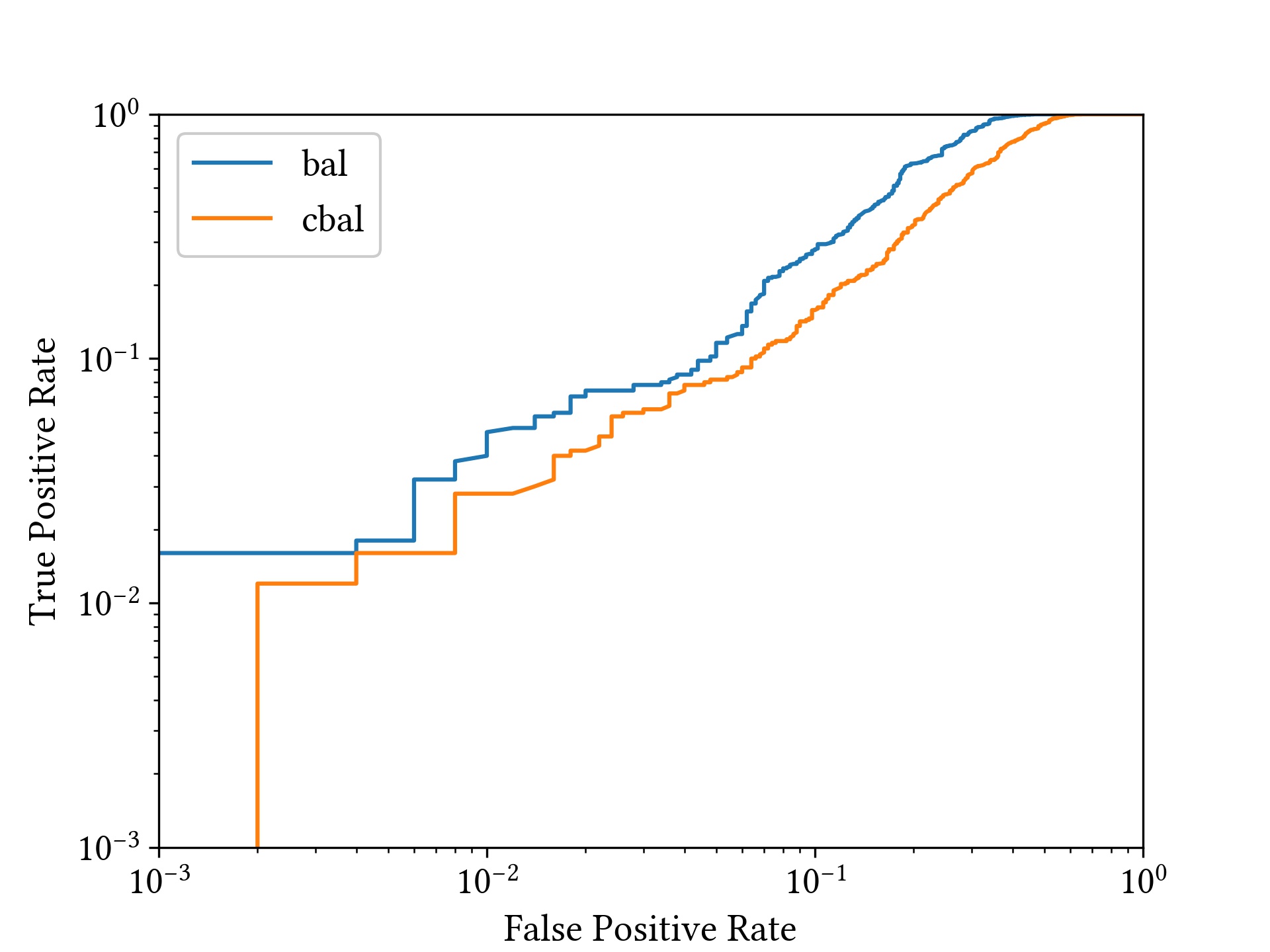}
		\caption*{(c)CIFAR10-M3} %?????subcaption????????????????????????????????????
		\label{pt:overfitting ROC(c)}%??????
  \end{minipage}
  \qquad
  \begin{minipage}{0.32\linewidth}
    \centering
    \includegraphics[width=0.9\linewidth]{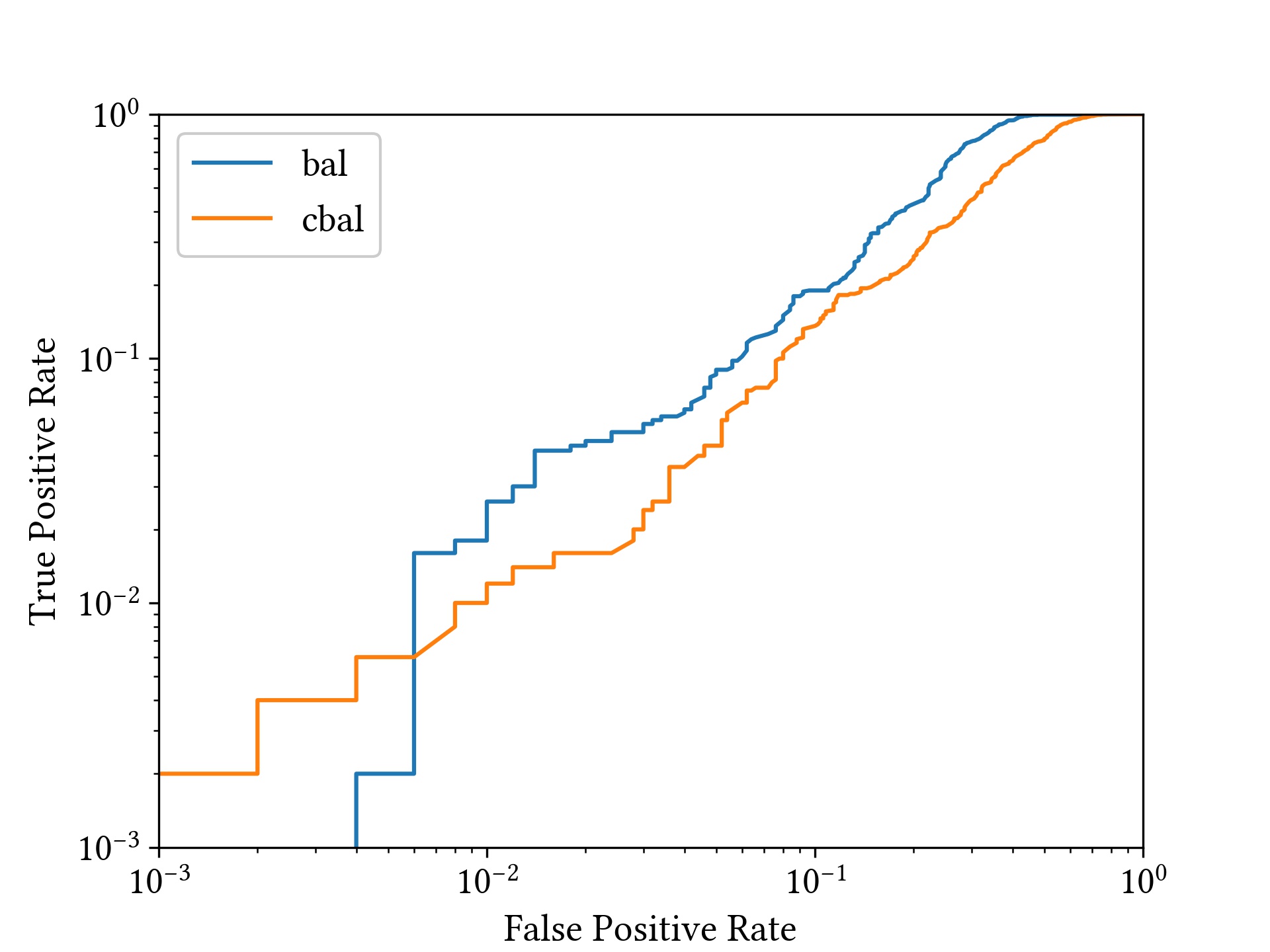}
		\caption*{(d)CIFAR100-M1} %?????subcaption????????????????????????????????????
		\label{pt:overfitting ROC(d)}%??????
  \end{minipage}
  \begin{minipage}{0.32\linewidth}
    \centering
    \includegraphics[width=0.9\linewidth]{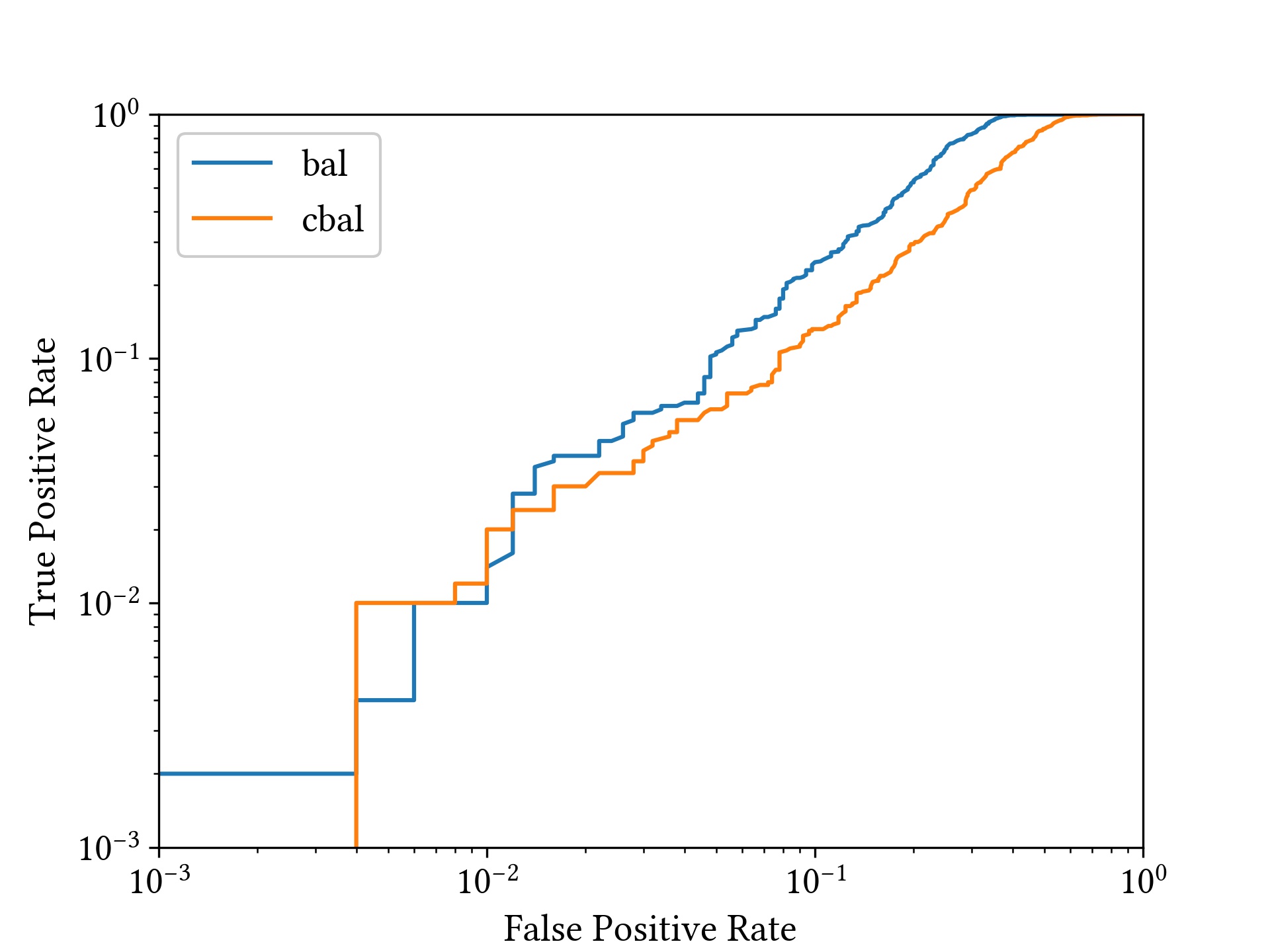}
		\caption*{(e)CIFAR100-M2} %?????subcaption????????????????????????????????????
		\label{pt:overfitting ROC(e)}%??????
  \end{minipage}
  \begin{minipage}{0.32\linewidth}
    \centering
    \includegraphics[width=0.9\linewidth]{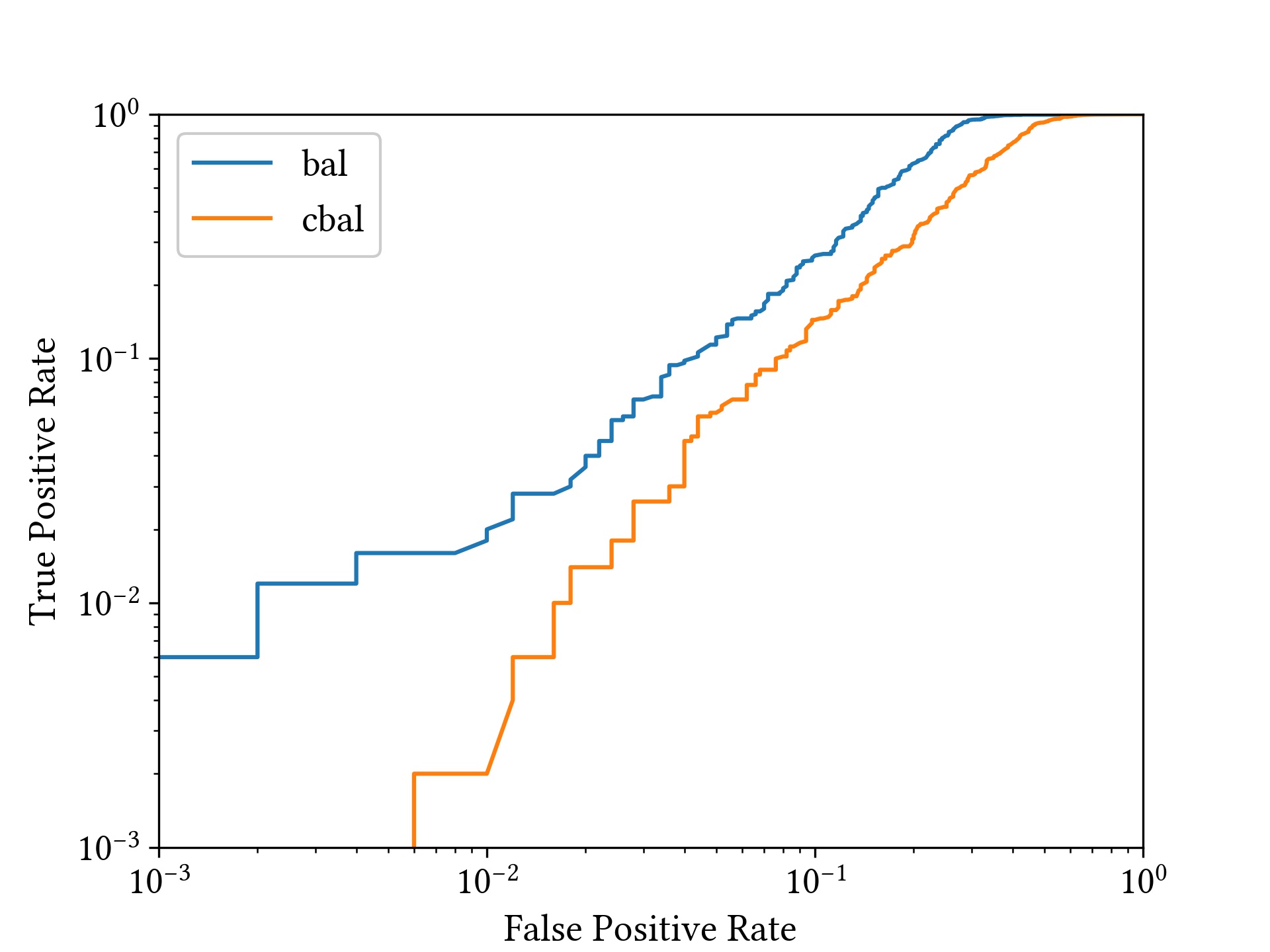}
		\caption*{(f)CIFAR100-M3} %?????subcaption????????????????????????????????????
		\label{pt:overfitting ROC(f)}%??????
  \end{minipage}
  \caption{Full log-scale ROC figures of untargeted MIAs under target models with different levels of overfitting.}
  \label{pt:overfitting ROC}%??????
  \Description{A full log-scale ROC figure in different overfitting.}
\end{figure*}

As Figure \ref{pt:overfitting ROC} shows, untargeted MIA achieves higher TPR at the same FPR in the balanced set compared to the cbalanced set in almost all models.
Table \ref{tab:overfitting with cbalanced} provides more detailed information about evaluation metrics such as TPR at 0.1\% FPR and TPR at 1\% FPR metrics.
When the evaluation sets are the same, the performance of these attacks on the balanced set tend to rise sharply with model overfitting, especially in the CIFAR10 dataset, where the attack's TPR at 0.1\% FPR metric is eight times higher in the M3 model than in the M1 model, while no obvious trend can be observed in the cbalanced set.
When the target models are the same, the evaluation metrics of untargeted MIA in the balanced set are much better than those in the cbalanced set.

Based on the above two findings, we offer a heuristic hypothesis that the untargeted MIA performs much worse in correctly identified samples than in misclassified samples.
A strong argument in favor of this hypothesis can be made by the observation that the evaluation results of the balanced set dramatically improve as the number of samples that are misclassified  rises due to the target model's increased overfitting.
However, too many misclassified samples in the balanced set may lead to an overestimation of the real power of untargeted MIA because realistic models are always well-trained to classify data correctly.

On the other hand, the cbalanced set suffers less from model overfitting and better captures the effectiveness of the untargeted HSJA because it does not contain any misclassified samples.
As a result, we think that the cbalanced set is more rigorous than the balanced set, more accurately represents how attacks perform in practical settings, and is less susceptible to model overfitting.
More explanations for our heuristic hypothesis can be found in the discussion.

\begin{table}[htp]
  \caption{Attack performance of untargeted MIA under different target models and evaluation datasets}
  \label{tab:overfitting with cbalanced}
  \begin{tabular}{ccccccc}
    \toprule
    \multirow{2}{*}{\begin{tabular}[c]{@{}c@{}}Target\\ Model\end{tabular}} & \multicolumn{2}{c}{TPR at 0.1\%FPR} & \multicolumn{2}{c}{TPR at 1\%FPR} & \multicolumn{2}{c}{AUC} \\ \cline{2-7} 
                                                                            & bal           & cbal         & bal          & cbal        & bal        & cbal       \\
    \midrule
    CIFAR10-M1                                                              & 0.002         & 0            & 0.016        & 0.008       & 0.698      & 0.629      \\
    CIFAR10-M2                                                              & 0.006         & 0.004        & 0.018        & 0.01        & 0.753      & 0.671      \\
    CIFAR10-M3                                                              & 0.016         & 0            & 0.038        & 0.028       & 0.822      & 0.727      \\ 
    \midrule
    CIFAR100-M1                                                             & 0             & 0.002        & 0.018        & 0.01        & 0.784      & 0.672      \\
    CIFAR100-M2                                                             & 0.002         & 0            & 0.01         & 0.012       & 0.809      & 0.693      \\
    CIFAR100-M3                                                             & 0.006         & 0            & 0.016        & 0.002       & 0.832      & 0.719      \\
    \bottomrule
    \end{tabular}
\end{table}

\begin{table}[htp]
  \caption{Target models with different degrees of overfitting}
  \label{tab:overfitting with accuracy}
  \begin{tabular}{lcc}
    \toprule
    Target Model        & Training Set Size & Train Test Gap \\
    \midrule
    CIFAR10-M1          & 50000             & 0.18           \\
    CIFAR10-M2          & 10000             & 0.24           \\
    CIFAR10-M3          & 5000              & 0.28           \\
    CIFAR100-M1         & 50000             & 0.34           \\
    CIFAR100-M2         & 40000             & 0.39           \\
    CIFAR100-M3         & 30000             & 0.42           \\
    \bottomrule
  \end{tabular}
\end{table}

\subsection{Decision Boundary Distance}
\label{subsec: decision boundary}
\begin{figure}[htp]
  \centering
  \includegraphics[width=\linewidth]{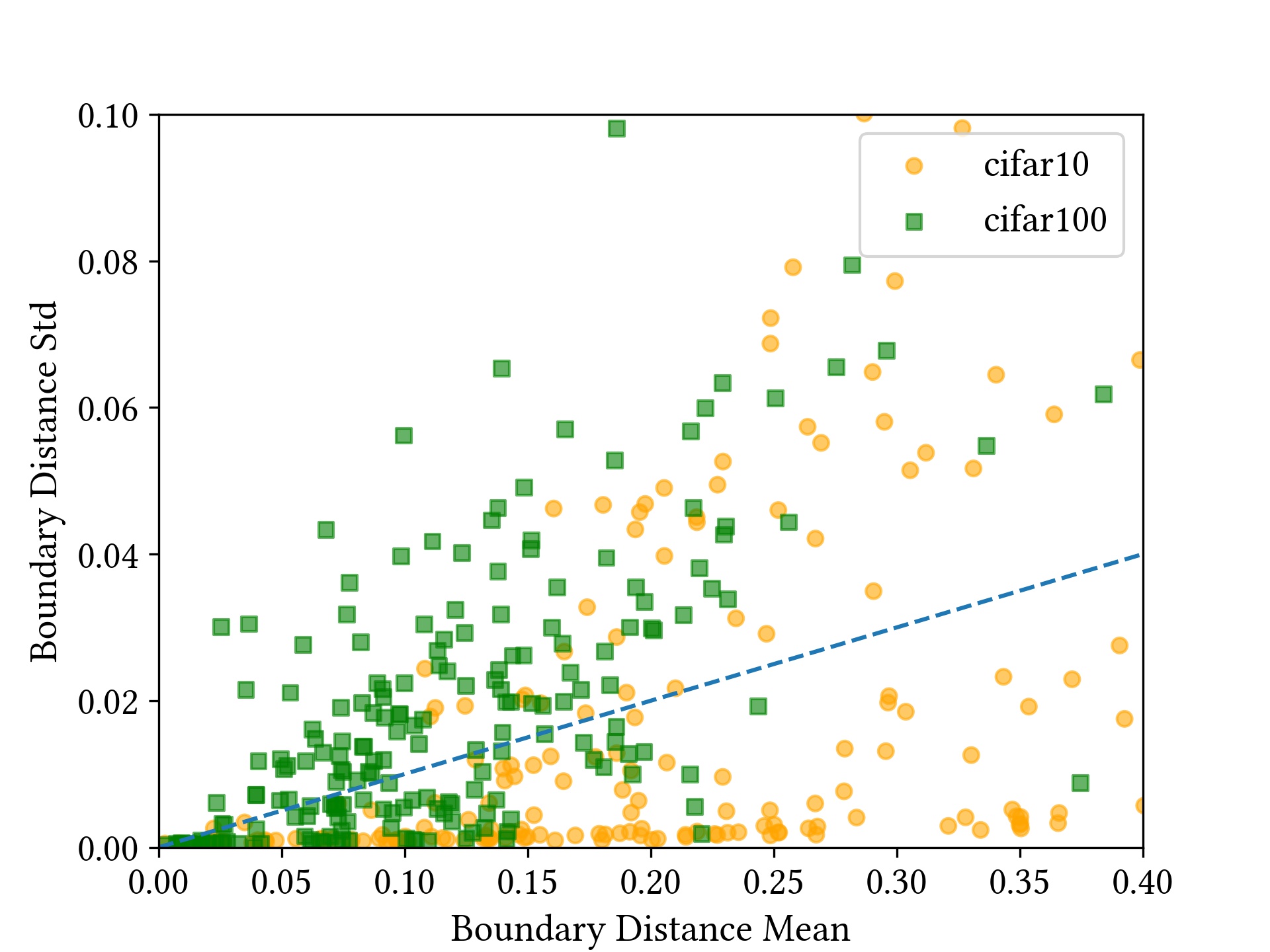}
  \caption{A scatter plot of the standard deviation of decision boundary distances versus the mean of decision boundary distances.}
  \label{pt:std_mean_scatter}
  \Description{A scatter.}
\end{figure}

In this section, we measure the stability of the decision boundary distance obtained by the three different HSJA algorithm and explore the key factor affecting this stability.
\paragraph{Causes of decision boundary distance instability}
In order to assess the stability, we performed untargeted HSJA ten times on a cbalanced set with a sample size of 200 and recorded the mean and standard deviation of the boundary distances for each sample, as shown in Figure \ref{pt:std_mean_scatter}.
We also defined a sample with a standard deviation $d_{std}$ one order of magnitude below the mean $d_{mean}$ as a stable sample and vice versa as a bias sample, respectively, correspond to the samples above and below the blue dashed line whose slope is 0.1 in the figure.

As Figure \ref{pt:std_mean_scatter} shows, a considerable number of samples fail to obtain a stable boundary distance with untargeted HSJA since their data points fall above the blue dashed line.
With additional data gathered from the 10 experiments, we discovered that samples with the same target class were more likely to acquire stable boundary distances.

As shown in Table \ref{tab:instability with label}, nearly all the samples with the same target class have stable boundary distances both in the CIFAR10 and CIFAR100 datasets.
Another interesting finding is that there are roughly the same numbers of samples that have diverse target classes but constant boundary distances across different datasets.
These two findings indicate that there is a close relationship between the target class and sample stability, and the difference of datasets mainly prompt the interconversion between stable samples of the same target class and deviating samples of different target classes.

\begin{figure}[htp]
  \centering
  \includegraphics[width=\linewidth]{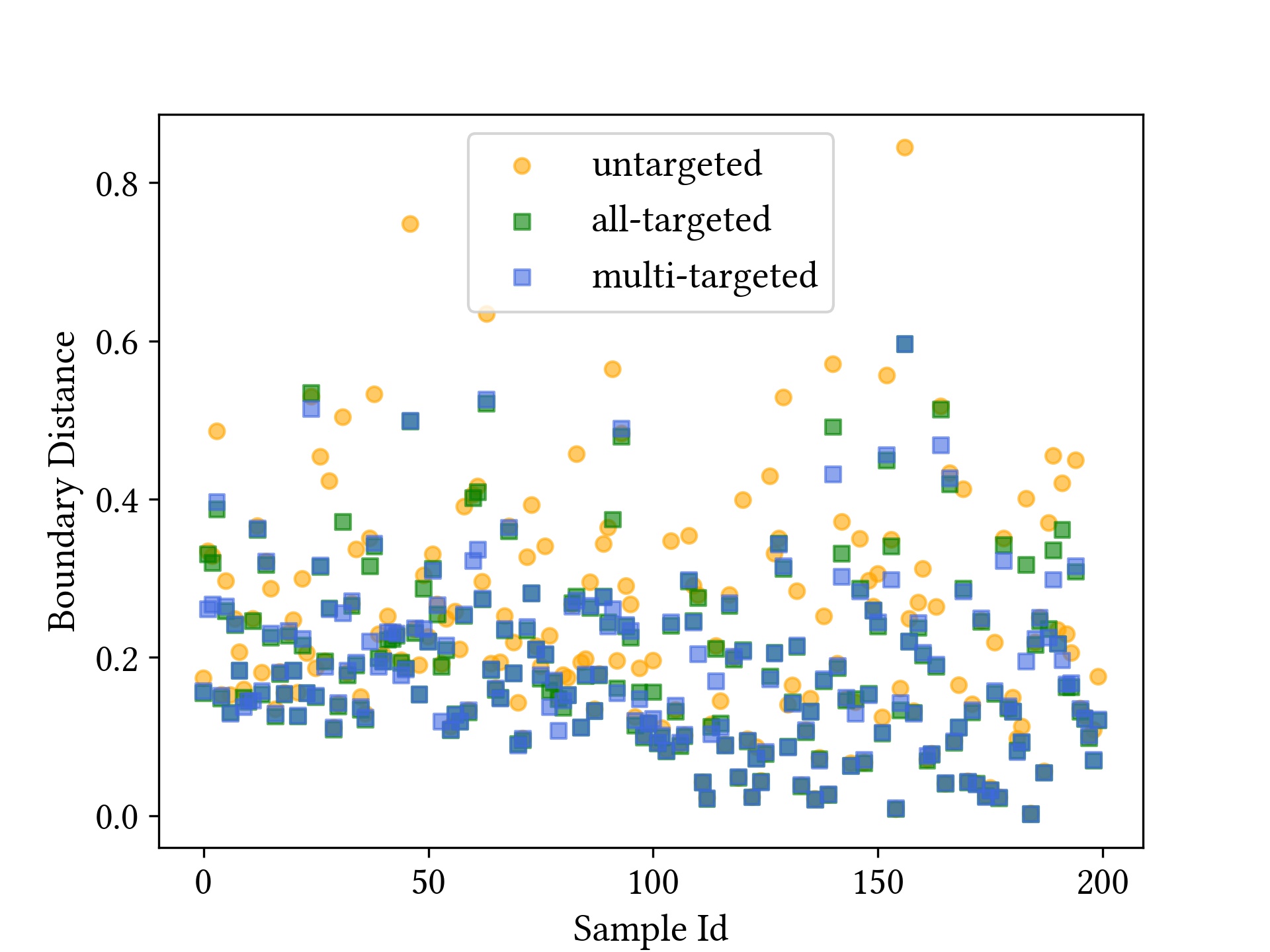}
  \caption{A scatter plot illustrates the decision boundary distances obtained for untargeted HSJA, all-targeted HSJA and multi-targeted HSJA at different sample points in CIFAR10 dataset.}
  \label{pt:cifar10_distance_sample_scatter}
  \Description{A scatter.}
\end{figure}

\begin{figure}[htp]
  \centering
  \includegraphics[width=\linewidth]{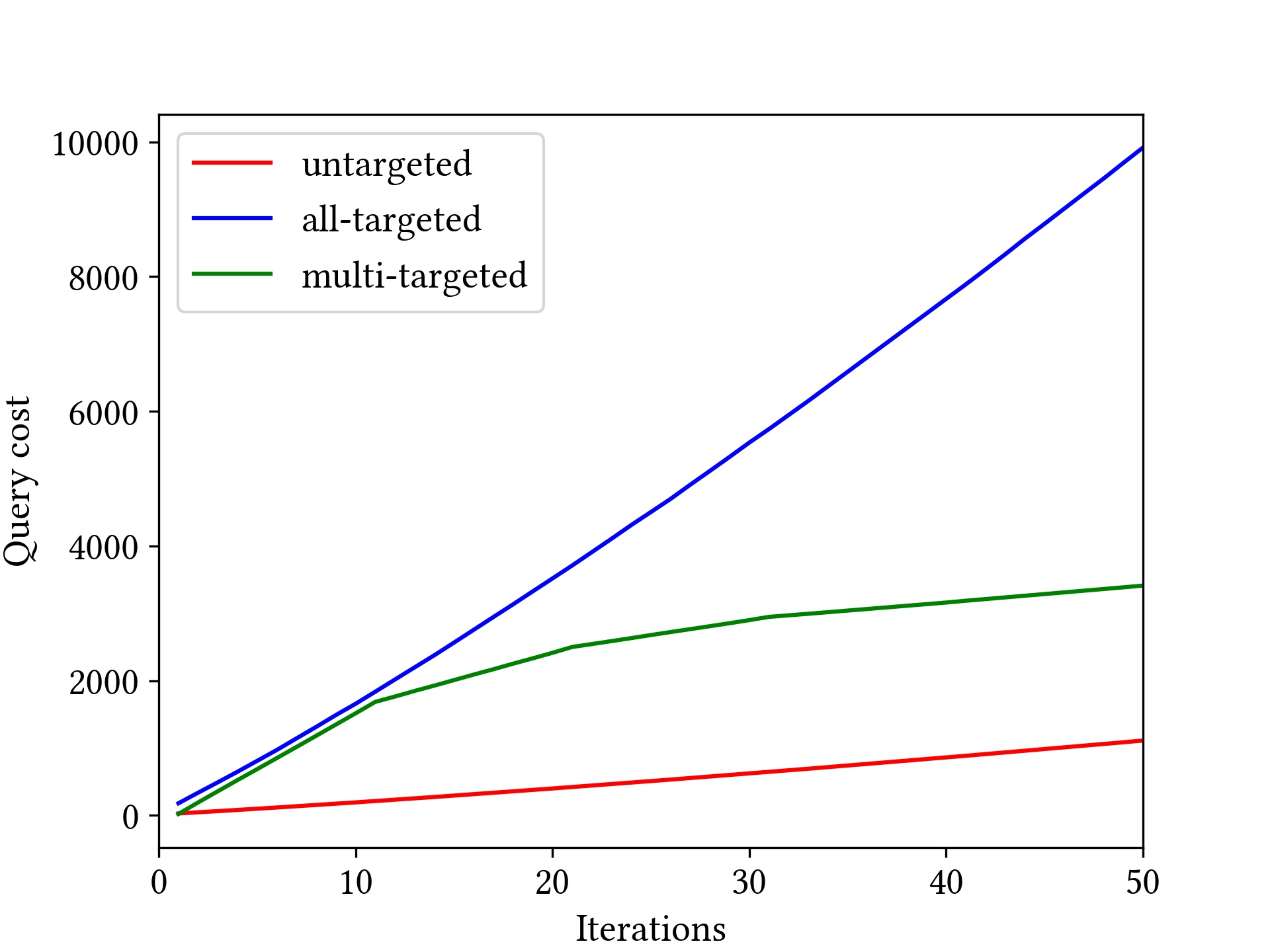}
  \caption{Query cost of three different HSJA with different number of iterations.}
  \label{pt:cifar10_query_with_iterations}
  \Description{A query.}
\end{figure}

\paragraph{Different HSJA}
Since the target class is the key factor determining the decision boundary distance, we propose all-targeted HSJA and multi-targeted HSJA to obtain stable minimum boundary distance by traversing target labels.
We compared the boundary distances acquired by three different HSJA algorithms, including untargeted HSJA, all-targeted HSJA and multi-targeted HSJA, to measure their efficiency.
For consistency and fairness's sake, all three methods were run on the same cbalanced set with one fixed correctly classified image serving as the initial adversarial sample for each target class.
As shown in Figure \ref{pt:cifar10_distance_sample_scatter}, the boundary distances obtained by untargeted HSJA are substantially larger than those obtained by all-targeted HSJA or multi-targeted HSJA in about half of the samples.

In order to further compare the gap between the all-targeted HSJA and the multi-targeted HSJA, the shortest boundary distanced obtained from these three algorithms is defined as $d_{min}$ and the minimum boundary distance region is defined as $[d_{min}, 1. 1*d_{min}]$.
All samples using all-targeted HSJA and nearly all samples using multi-targeted HSJA, as shown in Table \ref{tab:minimum boundary distancel}, reach the minimum boundary distance region, indicating that the minimum perturbed points generated by both approaches are comparable.

However, as the Figure \ref{pt:cifar10_query_with_iterations} illustrates, query costs of multi-targeted HSJA is much less than costs of all-targeted HSJA, representing a higher query efficiency.
Therefore, our multi-targeted HSJA outperforms the other two algorithms because it can obtain shorter boundary distances with fewer query cost.

\begin{table}[htp]
  \caption{The relationship between decision boundary target classes and decision boundary stability}
  \label{tab:instability with label}
  \begin{tabular}{ccccc}
    \toprule
    \multirow{2}{*}{\begin{tabular}[c]{@{}c@{}}Target\\ Label\end{tabular}} & \multicolumn{2}{c}{CIFAR10} & \multicolumn{2}{c}{CIFAR100} \\ \cline{2-5} 
                                                                            & stable    & bias  & stable  & bias       \\
    \midrule
    Same Label                                                              & 92        & 1       & 37       & 1       \\
    Not Same                                                                & 44        & 63      & 45       & 117        \\
    \bottomrule
    \end{tabular}
\end{table}

\begin{table}[htp]
  \caption{Number of samples reaching the minimum boundary distance region}
  \label{tab:minimum boundary distancel}
  \begin{tabular}{ccc}
    \toprule
    HSJA TYPE               & reached   & not reached\\ 
    \midrule
    Untargeted              & 107       & 93 \\
    Multi-targeted          & 199       & 1  \\
    All-targeted            & 200       & 0  \\
    \bottomrule
    \end{tabular}
\end{table}

\subsection{Membership Score}
\label{subsec:membership score}
In this section, we contrast how multi-class adaptive MIA attacks with relative boundary distance as the membership score perform in comparison to untargeted MIAs with single boundary distance as the membership score.
For the completeness of the comparison, we additionally consider the case where the boundary distance obtained from a multi-targeted HSJA and the relative boundary distance obtained from untargeted HSJA are used as membership scores, which are also referred to as multi-targeted MIAs and untargeted adaptive MIAs, respectively.

\begin{figure*}[htp]
  \centering
  \begin{minipage}{0.49\linewidth}
    \centering
    \includegraphics[width=0.9\linewidth]{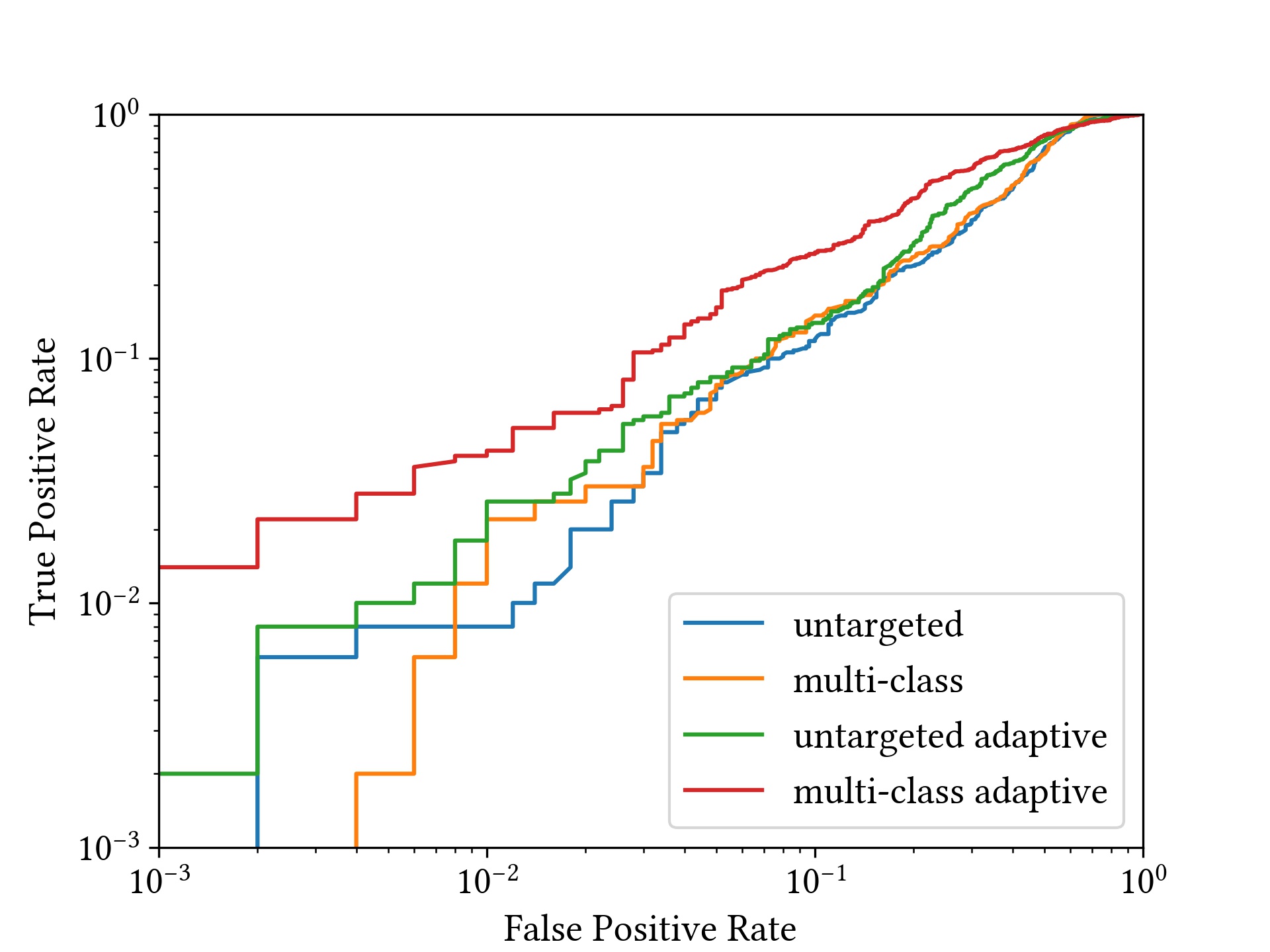}
    \subcaption{CIFAR10}
    \label{pt:distance_with_membership_score(a)}
  \end{minipage}
  \begin{minipage}{0.49\linewidth}
    \centering
    \includegraphics[width=0.9\linewidth]{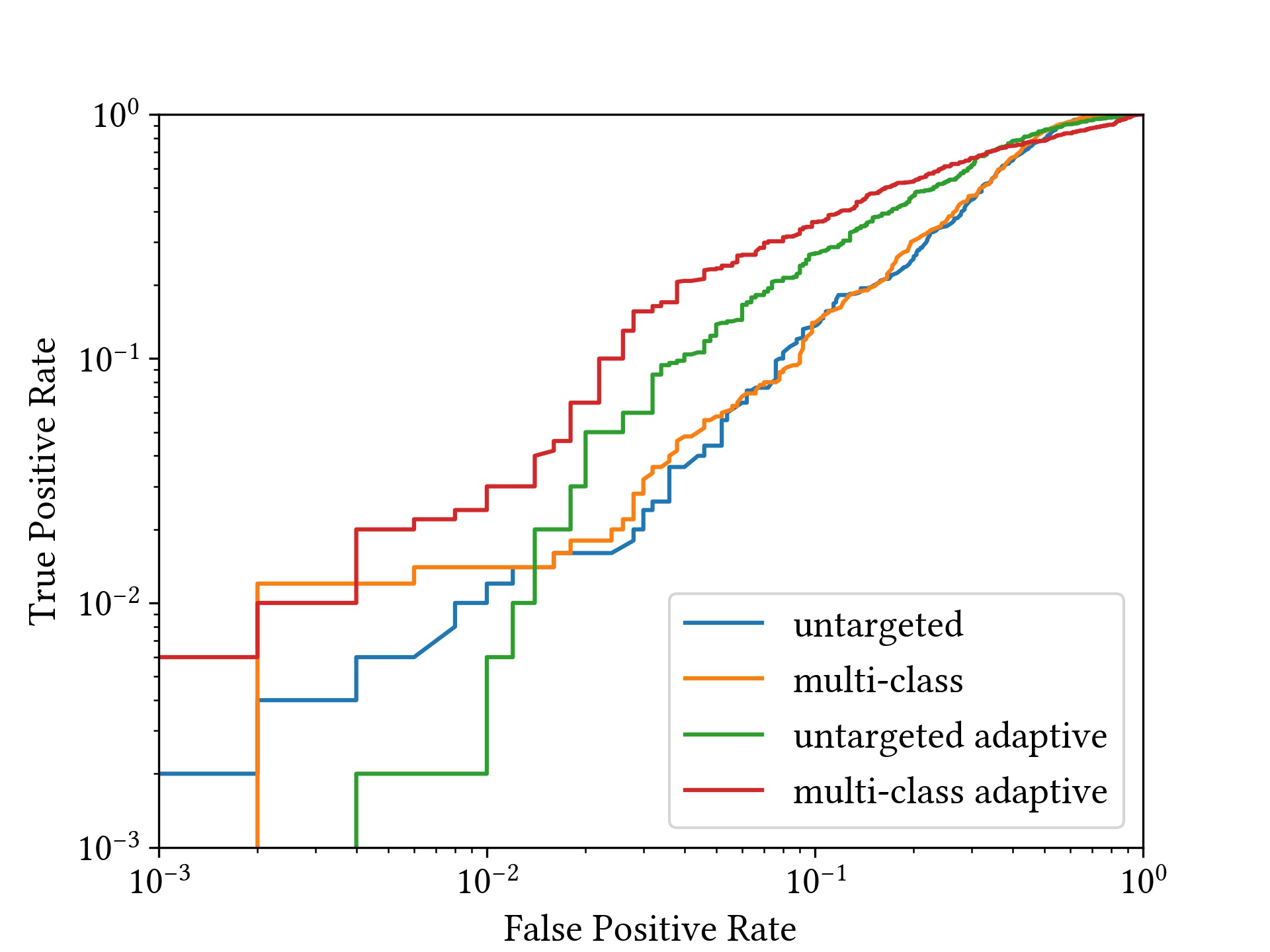}
    \subcaption{CIFAR100}
    \label{pt:distance_with_membership_score(b)}
  \end{minipage}
  \caption{Boundary distance with different membership score in different datasets.}
  \label{pt:distance_with_membership_score}
  \Description{A scatter.}
\end{figure*}

As shown in Figure \ref{pt:distance_with_membership_score}, our multi-class adaptive attack outperforms the other three attacks in almost the entire log-scale ROC curve, particularly for TPR at very low FPR.
Another noteworthy point is that multi-targeted attacks perform slightly better than untargeted attacks, indicating that the correct decision boundary distance assists in distinguishing between members and nonmembers.

\begin{table*}[h]
  \caption{MIA performance under different membership score options.}
  \label{tab:different membership score metrics}
  \begin{tabular}{ccccccc}
    \toprule
    \multirow{2}{*}{\begin{tabular}[c]{@{}c@{}}Membership\\ Score\end{tabular}} & \multicolumn{3}{c}{CIFAR10} & \multicolumn{3}{c}{CIFAR100} \\ \cline{2-7} 
                            & TPR at 0.1\%FPR    & TPR at 1\%FPR   & AUC & TPR at 0.1\%FPR    & TPR at 1\%FPR   & AUC\\ 
    \midrule
    Untargeted Distance                 & 0.0\%       & 0.8\%    & 0.629       & 0.2\%    & 1.0\%    & 0.672\\
    Multi-class Distance               & 0.0\%       & 1.2\%    & 0.638       & 0.0\%    & 1.4\%    & 0.681\\
    Untargeted adaptive Distance       & 0.2\%       & 1.8\%    & 0.665       & 0.0\%    & 0.2\%    & 0.729\\
    Multi-class adaptive Distance      & 1.4\%       & 4.0\%    & 0.715       & 0.6\%    & 2.4\%    & 0.723\\
    \bottomrule
    \end{tabular}
\end{table*}

Table \ref{tab:different membership score metrics} provides some more detailed metrics, and single decision boundary distance has been shown to be a poor option for membership score due to its extremely low TPR at 0.1\% FPR.
An interesting point is that untargeted adaptive MIA has a high AUC metric along with a low TPR at 0.1\% FPR metric.
A reasonable explanation is that although the adaptive membership scores increase the gap between the minimum boundary distance distributions of the member and non-member samples, the unstable boundary distances obtained by the untargeted HSJA prevent some samples from obtaining the minimum boundary distances.
In summary, our multi-class adaptive MIA uses the relative decision boundary distances obtained by multi-targeted HSJA as membership scores, which can effectively identify more member samples in a low error rate scenario.

\section{DISCUSSION}
In this section, we first go a step further by explaining why the untargeted MIA performed much better in incorrectly classified samples than in correctly classified ones.
Then, we further quantified the correlation between the decision boundary distance and the target classes using suitable metrics.
Finally, we discuss the key factors that make the multi-class adaptive MIA perform better than others.
\paragraph{Evaluation set}
In section \ref{subsec:evaluation set}, we proposed a hypothesis that the traditional label-only MIA performs much worse on correctly classified samples than on misclassified samples, and we provide an explanation here.
To the best of our knowledge, the earliest label-only MIA\cite{yeom2018privacy} assumed that the ground truth of all samples could be obtained manually, and each member sample was identified by comparing the hard labels returned by the model with the ground truth.
The intuition behind this attack was that a well-trained model would always correctly classify members of the training set, and only nonmembers were likely to be misclassified.

Although the later untargeted MIAs employed boundary distance as the new membership score instead, this baseline attack was kept as a prior due to its simplicity and efficiency.
As a result, their attack performance on the balanced set was exaggerated because evaluation results were essentially the outcomes of hybrid attacks.
In contrast, the evaluation results of cbalanced sets honestly reflect the effectiveness of their claimed attack strategies because misclassified samples have been removed.

Instead of criticizing this hybrid attack method or balanced sets, our goal of establishing cbalanced sets is to more thoroughly assess the attack strategy's capacity to discriminate between members and nonmembers.
\paragraph{Minimum boundary distance}

\begin{figure*}[htp]
  \centering
  \begin{minipage}{0.32\linewidth}
    \centering
    \includegraphics[width=0.9\linewidth]{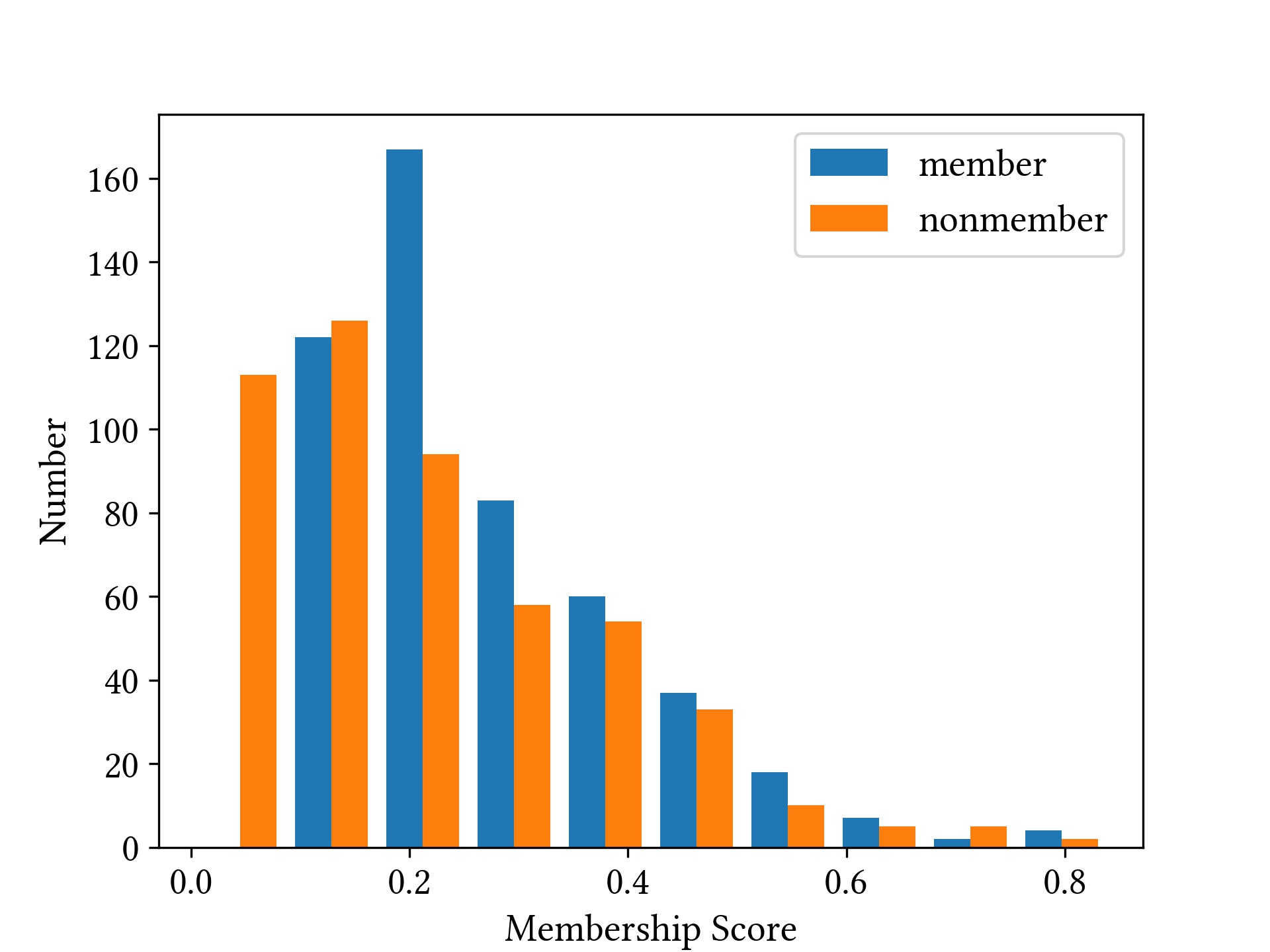}
    \subcaption{Untargeted}
    \label{pt:boundary_distance_histogram(a)}
  \end{minipage}
  \begin{minipage}{0.32\linewidth}
    \centering
    \includegraphics[width=0.9\linewidth]{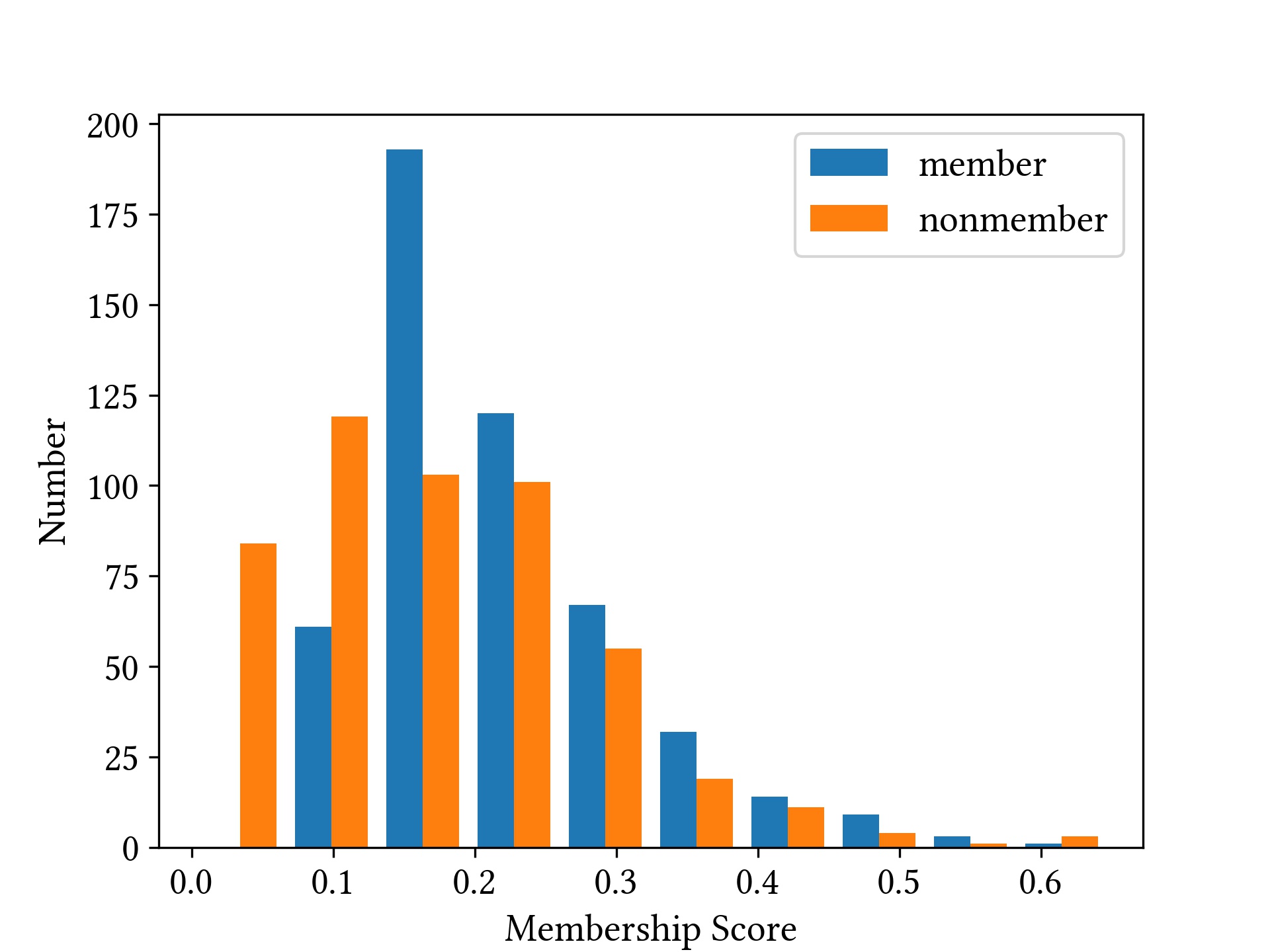}
    \subcaption{Multi-targeted}
    \label{pt:boundary_distance_histogram(b)}
  \end{minipage}
  \begin{minipage}{0.32\linewidth}
    \centering
    \includegraphics[width=0.9\linewidth]{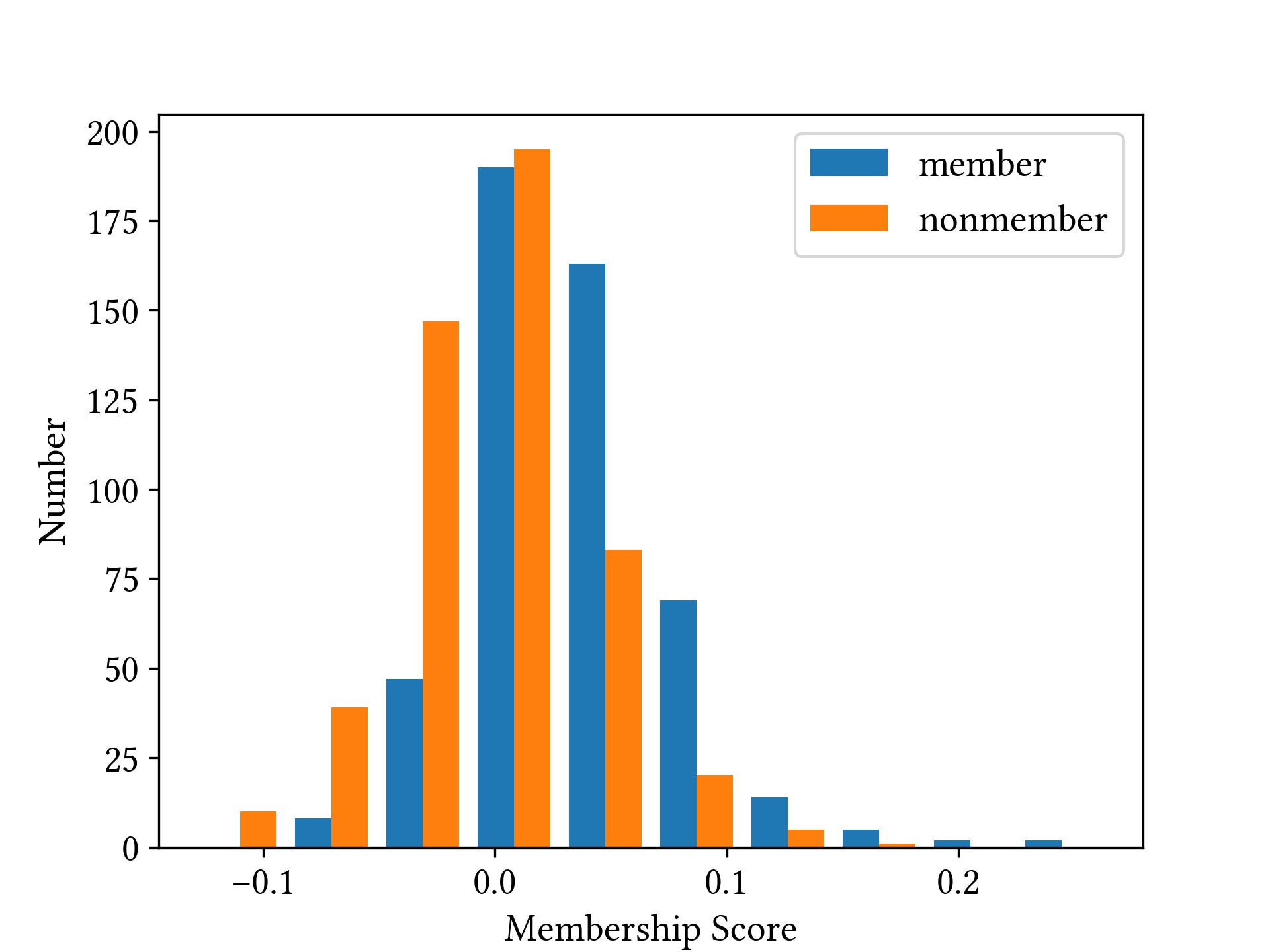}
    \subcaption{Multi-class adaptive}
    \label{pt:boundary_distance_histogram(c)}
  \end{minipage}
  \caption{Histograms of different membership scores in CIFAR10 dataset.}
  \label{pt:boundary_distance_histogram}
  \Description{A.}
\end{figure*}

In section \ref{subsec:evaluation set}, we found an underlying relationship between the boundary distance stability and the target class, and here we use relative coefficients to analyze the relationship.
Since both boundary distance stability and target classes are discrete statistical variables, we utilize the Spearman correlation coefficient to quantify the degree of their association.

The correlation coefficients for CIFAR10 and CIFAR100, respectively, are 0.618 and 0.555, indicating a positive correlation between decision boundary distance stability and the same target class.
The related p-values are 1.85e-22 and 1.47e-17, respectively, representing the positive correlation is obvious.
% This association also explains why the untargeted HSJA has difficulties reaching the minimum decision boundary distance, i.e., the random initial images may have distinct target classes and ultimately result in variable boundary distances.

Vo et al. \cite{vo2021ramboattack} also mentioned the difficulty of obtaining the minimum boundary distance with untargeted HSJA and attributed it to the selection of the initial images.
Unlike their solution of using switching points to geometrically bypass local minima, we traversed all classes to ensure that the target class with the minimum boundary distance is selected.

\paragraph{Membership score}
In section \ref{subsec:membership score}, the single decision boundary distance failed to distinguish between members and nonmembers at a low FPR and the ignored sample diversity may be the main reason for this failure.
As shown in Figure \ref{pt:boundary_distance_histogram(a)}, although the member sample has a higher mean membership score than the nonmember sample, there is still a large overlap in the range of their distributions, which makes it difficult to identify low FPR membership samples under a fixed threshold.
Relative boundary distances, on the other hand, are equivalent to creating a unique adaptive threshold for each sample when used as a membership score. 
Even for nonmember samples with long boundary distance, a short relative average distance can be achieved by subtracting the long average boundary distances of neighboring points.
Figure \ref{pt:boundary_distance_histogram(b)} illustrates how the relative distance distribution of the member sample is farther away from that of the non-member sample, allowing for a high TPR at a low FPR.

% \begin{figure}[h]
%   \centering
%   \includegraphics[width=\linewidth]{fig/compare_untar_histogram.jpg}
%   \caption{boundary distance as membership score histogram.}
%   \label{pt:cifar10_boundary_distance_membership_score_histogram}
%   \Description{A cifar10_boundary_distance_membership_score_histogram.}
% \end{figure}

% \begin{figure}[h]
%   \centering
%   \includegraphics[width=\linewidth]{fig/compare_adaptive_histogram.jpg}
%   \caption{boundary distance with different membership score.}
%   \label{pt:cifar10_relative_boundary_distance_membership_score_histogram}
%   \Description{A cifar10_relative_boundary_distance_membership_score_histogram.}
% \end{figure}

\section{RELATED WORK}
Membership inference attacks have gained widespread attention as one of the simplest forms of information leakage today.
In this section, we focus on the areas that are most relevant to our research.
\paragraph{Score-based membership inference attack}
Shokri et al. \cite{shokri2017membership} proposed the first membership inference attack to infer whether unknown samples are members from the training set.
They trained multiple shadow models using the same distribution as the training set but disjoint auxiliary information to mimic the performance of the target model.
These shadow models provided adversaries with an unlimited number of queries to train multiple attack models to identify the member samples.
Salem et al. \cite{salem2019ml} simplified this process by training only one shadow model and achieved attack performance similar to that of multiple shadow models.
Song et al. \cite{song2021systematic} argued that the attack model contained strong randomness and proposed a benchmark attack without the attack model based on a modified predictive entropy estimate.
Hui et al. \cite{hui2021practical} additionally considered the case of unknown ground truth and proposed an attack method based on differential comparison.

However, each of the above MIAs makes the presumption that the adversary has at least one prior, namely the sample's confidence value, which is challenging to satisfy in the real world.
As a result, more research has recently concentrated on applying MIA in more practical scenarios, such as the label-only setting, where the target model only provides a predicted label rather than the confidence value.

\paragraph{Label-only membership inference attack}
Yeom et al. \cite{yeom2018privacy} proposed the first label-only MIA that classified samples as member samples or vice versa depending on whether predicted labels and ground truth were equal.
Although this strategy outperformed random guesses on misclassified samples, it fell short on correctly classified samples.
Choquette et al. \cite{choquette2021label} developed a decision-based inference attack to address this weakness by using the sample's robustness as a substitute for the confidence value.
The decision boundary distance obtained by the decision-based black-box adversarial attack algorithm HopSkipJump was used to measure how robust the sample was.
In addition, Li et al. \cite{li2021membership}suggested a universal threshold-choosing approach for this attack in various datasets.
Zhang et al. \cite{zhang2022label} extended the target model from classification models to semantic segmentation models.
Lu et al. \cite{lu2022label} applied label-only membership inference attacks into machine unlearning.

Apart from minimizing prior knowledge of MIAs, it was discovered that these attacks' assessment mechanisms, which used average-case evaluation metrics, did not accurately reflect their attack performance.
\paragraph{Non-average evaluation}
In order to construct and evaluate the performance of attack strategies, most attacks follow a membership inference game formulation\cite{yeom2018privacy} that emphasizes membership relation inference in a balanced set of equal number of members and nonmembers.
Therefore, average-case metrics (e.g. balanced accuracy and AUC) were frequently used to quantify the performance of an attack strategy.
Rezaei et al. \cite{rezaei2021difficulty}argued that these average-case metrics frequently misidentified nonmembers as members due to their high FPR, which is unreported.
In contrast, TPR at low FPR metrics and log-scale ROC curves have been viewed as better metrics since they exhibit the quantity of member samples detected with a very low error rate, in line with the goals of MIA.
In addition, Ye et al. \cite{ye2022enhanced} provided an innovative template for inference games and an explanation of the success rate of attacks on various game instances.

Most of the old methods were abandoned as a result of examination using these new measurements, but many new and effective methods also emerged.
Watson et al. \cite{watsonimportance} proposed difficulty calibration that took into account the difficulty of correctly classifying the target sample, effectively reducing the FPR value.
Carlini et al. \cite{carlini2022membership} obtained a high TPR at low FPR by sampling multiple shadow models for each sample and calculating the likelihood ratio instead of a fixed threshold.
Rezaei et al. \cite{rezaei2022efficient} sampled in a subspace of each sample instead of the sample under different shadow models, maintaining high performance while reducing computational costs.

\section{CONCLUSION}
In this paper, we introduce multi-class adaptive MIA, which obtains the decision boundary distance with our multi-targeted HSJA and uses relative decision boundary distances as membership scores to distinguish between members and nonmembers.
Numerous experiments showed that our multi-targeted HSJA was able to obtain the shortest decision boundary distance with a limited number of queries and that our multi-class adaptive MIA outperformed the untargeted MIAs across the whole log-scale ROC curve.
Additionally, we found a significant difference between the performance of the well-known label-only attack on incorrectly and correctly classified samples, and we recommended using the cbalanced set as the evaluation set rather than the balanced set.
We expect that by more quickly traversing the target class's decision boundaries, geometry-based decision attacks will help future work further reduce the time and query count costs.
\begin{acks}
  This work was supported by National Natural Science Foundation of China under grant number 72274138.
\end{acks}

% \section{Appendices}

% If your work needs an appendix, add it before the
% ``\verb|\end{document}|'' command at the conclusion of your source
% document.

% Start the appendix with the ``\verb|appendix|'' command:
% \begin{verbatim}
%   \appendix
% \end{verbatim}
% and note that in the appendix, sections are lettered, not
% numbered. This document has two appendices, demonstrating the section
% and subsection identification method.

\bibliographystyle{ACM-Reference-Format}
\bibliography{sample-base}


\begin{thebibliography}{36}

%%% the Web address.

\ifx \showCODEN    \undefined \def \showCODEN     #1{\unskip}     \fi
\ifx \showDOI      \undefined \def \showDOI       #1{#1}\fi
\ifx \showISBNx    \undefined \def \showISBNx     #1{\unskip}     \fi
\ifx \showISBNxiii \undefined \def \showISBNxiii  #1{\unskip}     \fi
\ifx \showISSN     \undefined \def \showISSN      #1{\unskip}     \fi
\ifx \showLCCN     \undefined \def \showLCCN      #1{\unskip}     \fi
\ifx \shownote     \undefined \def \shownote      #1{#1}          \fi
\ifx \showarticletitle \undefined \def \showarticletitle #1{#1}   \fi
\ifx \showURL      \undefined \def \showURL       {\relax}        \fi
% invisible to TeX
\providecommand\bibfield[2]{#2}
\providecommand\bibinfo[2]{#2}
\providecommand\natexlab[1]{#1}
\providecommand\showeprint[2][]{arXiv:#2}

\bibitem[Balle et~al\mbox{.}(2022)]%
        {balle2022reconstructing}
\bibfield{author}{\bibinfo{person}{Borja Balle}, \bibinfo{person}{Giovanni
  Cherubin}, {and} \bibinfo{person}{Jamie Hayes}.}
  \bibinfo{year}{2022}\natexlab{}.
\newblock \showarticletitle{Reconstructing training data with informed
  adversaries}. In \bibinfo{booktitle}{\emph{2022 IEEE Symposium on Security
  and Privacy (SP)}}. IEEE, \bibinfo{pages}{1138--1156}.
\newblock


\bibitem[Beltran et~al\mbox{.}(2021)]%
        {beltran2021male}
\bibfield{author}{\bibinfo{person}{Javier Beltran}, \bibinfo{person}{Aina
  Gallego}, \bibinfo{person}{Alba Huidobro}, \bibinfo{person}{Enrique Romero},
  {and} \bibinfo{person}{Llu{\'\i}s Padr{\'o}}.}
  \bibinfo{year}{2021}\natexlab{}.
\newblock \showarticletitle{Male and female politicians on Twitter: A machine
  learning approach}.
\newblock \bibinfo{journal}{\emph{European Journal of Political Research}}
  \bibinfo{volume}{60}, \bibinfo{number}{1} (\bibinfo{year}{2021}),
  \bibinfo{pages}{239--251}.
\newblock


\bibitem[Brendel et~al\mbox{.}(2017)]%
        {brendel2017decision}
\bibfield{author}{\bibinfo{person}{Wieland Brendel}, \bibinfo{person}{Jonas
  Rauber}, {and} \bibinfo{person}{Matthias Bethge}.}
  \bibinfo{year}{2017}\natexlab{}.
\newblock \showarticletitle{Decision-based adversarial attacks: Reliable
  attacks against black-box machine learning models}.
\newblock \bibinfo{journal}{\emph{arXiv preprint arXiv:1712.04248}}
  (\bibinfo{year}{2017}).
\newblock


\bibitem[Cardaioli et~al\mbox{.}(2020)]%
        {cardaioli2020predicting}
\bibfield{author}{\bibinfo{person}{Matteo Cardaioli}, \bibinfo{person}{Pallavi
  Kaliyar}, \bibinfo{person}{Pasquale Capuozzo}, \bibinfo{person}{Mauro Conti},
  \bibinfo{person}{Giuseppe Sartori}, {and} \bibinfo{person}{Merylin Monaro}.}
  \bibinfo{year}{2020}\natexlab{}.
\newblock \showarticletitle{Predicting Twitter users' political orientation: an
  application to the italian political scenario}. In
  \bibinfo{booktitle}{\emph{2020 IEEE/ACM International Conference on Advances
  in Social Networks Analysis and Mining (ASONAM)}}. IEEE,
  \bibinfo{pages}{159--165}.
\newblock


\bibitem[Carlini et~al\mbox{.}(2022)]%
        {carlini2022membership}
\bibfield{author}{\bibinfo{person}{Nicholas Carlini}, \bibinfo{person}{Steve
  Chien}, \bibinfo{person}{Milad Nasr}, \bibinfo{person}{Shuang Song},
  \bibinfo{person}{Andreas Terzis}, {and} \bibinfo{person}{Florian Tramer}.}
  \bibinfo{year}{2022}\natexlab{}.
\newblock \showarticletitle{Membership inference attacks from first
  principles}. In \bibinfo{booktitle}{\emph{2022 IEEE Symposium on Security and
  Privacy (SP)}}. IEEE, \bibinfo{pages}{1897--1914}.
\newblock


\bibitem[Chen et~al\mbox{.}(2020)]%
        {chen2020hopskipjumpattack}
\bibfield{author}{\bibinfo{person}{Jianbo Chen}, \bibinfo{person}{Michael~I
  Jordan}, {and} \bibinfo{person}{Martin~J Wainwright}.}
  \bibinfo{year}{2020}\natexlab{}.
\newblock \showarticletitle{Hopskipjumpattack: A query-efficient decision-based
  attack}. In \bibinfo{booktitle}{\emph{2020 ieee symposium on security and
  privacy (sp)}}. IEEE, \bibinfo{pages}{1277--1294}.
\newblock


\bibitem[Choquette-Choo et~al\mbox{.}(2021)]%
        {choquette2021label}
\bibfield{author}{\bibinfo{person}{Christopher~A Choquette-Choo},
  \bibinfo{person}{Florian Tramer}, \bibinfo{person}{Nicholas Carlini}, {and}
  \bibinfo{person}{Nicolas Papernot}.} \bibinfo{year}{2021}\natexlab{}.
\newblock \showarticletitle{Label-only membership inference attacks}. In
  \bibinfo{booktitle}{\emph{International conference on machine learning}}.
  PMLR, \bibinfo{pages}{1964--1974}.
\newblock


\bibitem[Del~Grosso et~al\mbox{.}(2022)]%
        {del2022leveraging}
\bibfield{author}{\bibinfo{person}{Ganesh Del~Grosso}, \bibinfo{person}{Hamid
  Jalalzai}, \bibinfo{person}{Georg Pichler}, \bibinfo{person}{Catuscia
  Palamidessi}, {and} \bibinfo{person}{Pablo Piantanida}.}
  \bibinfo{year}{2022}\natexlab{}.
\newblock \showarticletitle{Leveraging adversarial examples to quantify
  membership information leakage}. In \bibinfo{booktitle}{\emph{Proceedings of
  the IEEE/CVF Conference on Computer Vision and Pattern Recognition}}.
  \bibinfo{pages}{10399--10409}.
\newblock


\bibitem[Dwork(2006)]%
        {dwork2006differential}
\bibfield{author}{\bibinfo{person}{Cynthia Dwork}.}
  \bibinfo{year}{2006}\natexlab{}.
\newblock \showarticletitle{Differential privacy}. In
  \bibinfo{booktitle}{\emph{Automata, Languages and Programming: 33rd
  International Colloquium, ICALP 2006, Venice, Italy, July 10-14, 2006,
  Proceedings, Part II 33}}. Springer, \bibinfo{pages}{1--12}.
\newblock


\bibitem[Fernando et~al\mbox{.}(2021)]%
        {fernando2021deep}
\bibfield{author}{\bibinfo{person}{Tharindu Fernando},
  \bibinfo{person}{Harshala Gammulle}, \bibinfo{person}{Simon Denman},
  \bibinfo{person}{Sridha Sridharan}, {and} \bibinfo{person}{Clinton Fookes}.}
  \bibinfo{year}{2021}\natexlab{}.
\newblock \showarticletitle{Deep learning for medical anomaly detection--a
  survey}.
\newblock \bibinfo{journal}{\emph{ACM Computing Surveys (CSUR)}}
  \bibinfo{volume}{54}, \bibinfo{number}{7} (\bibinfo{year}{2021}),
  \bibinfo{pages}{1--37}.
\newblock


\bibitem[He et~al\mbox{.}(2016)]%
        {he2016deep}
\bibfield{author}{\bibinfo{person}{Kaiming He}, \bibinfo{person}{Xiangyu
  Zhang}, \bibinfo{person}{Shaoqing Ren}, {and} \bibinfo{person}{Jian Sun}.}
  \bibinfo{year}{2016}\natexlab{}.
\newblock \showarticletitle{Deep residual learning for image recognition}. In
  \bibinfo{booktitle}{\emph{Proceedings of the IEEE conference on computer
  vision and pattern recognition}}. \bibinfo{pages}{770--778}.
\newblock


\bibitem[Hui et~al\mbox{.}(2021)]%
        {hui2021practical}
\bibfield{author}{\bibinfo{person}{Bo Hui}, \bibinfo{person}{Yuchen Yang},
  \bibinfo{person}{Haolin Yuan}, \bibinfo{person}{Philippe Burlina},
  \bibinfo{person}{Neil~Zhenqiang Gong}, {and} \bibinfo{person}{Yinzhi Cao}.}
  \bibinfo{year}{2021}\natexlab{}.
\newblock \showarticletitle{Practical Blind Membership Inference Attack via
  Differential Comparisons}. In \bibinfo{booktitle}{\emph{ISOC Network and
  Distributed System Security Symposium (NDSS)}}.
\newblock


\bibitem[Jayaraman et~al\mbox{.}(2021)]%
        {jayaraman2021revisiting}
\bibfield{author}{\bibinfo{person}{Bargav Jayaraman}, \bibinfo{person}{Lingxiao
  Wang}, \bibinfo{person}{Katherine Knipmeyer}, \bibinfo{person}{Quanquan Gu},
  {and} \bibinfo{person}{David Evans}.} \bibinfo{year}{2021}\natexlab{}.
\newblock \showarticletitle{Revisiting Membership Inference Under Realistic
  Assumptions}.
\newblock \bibinfo{journal}{\emph{Proceedings on Privacy Enhancing
  Technologies}} \bibinfo{volume}{2021}, \bibinfo{number}{2}
  (\bibinfo{year}{2021}).
\newblock


\bibitem[Jia et~al\mbox{.}(2019)]%
        {jia2019memguard}
\bibfield{author}{\bibinfo{person}{Jinyuan Jia}, \bibinfo{person}{Ahmed Salem},
  \bibinfo{person}{Michael Backes}, \bibinfo{person}{Yang Zhang}, {and}
  \bibinfo{person}{Neil~Zhenqiang Gong}.} \bibinfo{year}{2019}\natexlab{}.
\newblock \showarticletitle{Memguard: Defending against black-box membership
  inference attacks via adversarial examples}. In
  \bibinfo{booktitle}{\emph{Proceedings of the 2019 ACM SIGSAC conference on
  computer and communications security}}. \bibinfo{pages}{259--274}.
\newblock


\bibitem[Krizhevsky et~al\mbox{.}(2009)]%
        {krizhevsky2009learning}
\bibfield{author}{\bibinfo{person}{Alex Krizhevsky}, \bibinfo{person}{Geoffrey
  Hinton}, {et~al\mbox{.}}} \bibinfo{year}{2009}\natexlab{}.
\newblock \showarticletitle{Learning multiple layers of features from tiny
  images}.
\newblock  (\bibinfo{year}{2009}).
\newblock


\bibitem[Li et~al\mbox{.}(2020)]%
        {li2020qeba}
\bibfield{author}{\bibinfo{person}{Huichen Li}, \bibinfo{person}{Xiaojun Xu},
  \bibinfo{person}{Xiaolu Zhang}, \bibinfo{person}{Shuang Yang}, {and}
  \bibinfo{person}{Bo Li}.} \bibinfo{year}{2020}\natexlab{}.
\newblock \showarticletitle{Qeba: Query-efficient boundary-based blackbox
  attack}. In \bibinfo{booktitle}{\emph{Proceedings of the IEEE/CVF conference
  on computer vision and pattern recognition}}. \bibinfo{pages}{1221--1230}.
\newblock


\bibitem[Li and Zhang(2021)]%
        {li2021membership}
\bibfield{author}{\bibinfo{person}{Zheng Li} {and} \bibinfo{person}{Yang
  Zhang}.} \bibinfo{year}{2021}\natexlab{}.
\newblock \showarticletitle{Membership leakage in label-only exposures}. In
  \bibinfo{booktitle}{\emph{Proceedings of the 2021 ACM SIGSAC Conference on
  Computer and Communications Security}}. \bibinfo{pages}{880--895}.
\newblock


\bibitem[Liu et~al\mbox{.}(2022a)]%
        {liu2022ml}
\bibfield{author}{\bibinfo{person}{Yugeng Liu}, \bibinfo{person}{Rui Wen},
  \bibinfo{person}{Xinlei He}, \bibinfo{person}{Ahmed Salem},
  \bibinfo{person}{Zhikun Zhang}, \bibinfo{person}{Michael Backes},
  \bibinfo{person}{Emiliano De~Cristofaro}, \bibinfo{person}{Mario Fritz},
  {and} \bibinfo{person}{Yang Zhang}.} \bibinfo{year}{2022}\natexlab{a}.
\newblock \showarticletitle{$\{$ML-Doctor$\}$: Holistic Risk Assessment of
  Inference Attacks Against Machine Learning Models}. In
  \bibinfo{booktitle}{\emph{31st USENIX Security Symposium (USENIX Security
  22)}}. \bibinfo{pages}{4525--4542}.
\newblock


\bibitem[Liu et~al\mbox{.}(2022b)]%
        {liu2022membership}
\bibfield{author}{\bibinfo{person}{Yiyong Liu}, \bibinfo{person}{Zhengyu Zhao},
  \bibinfo{person}{Michael Backes}, {and} \bibinfo{person}{Yang Zhang}.}
  \bibinfo{year}{2022}\natexlab{b}.
\newblock \showarticletitle{Membership inference attacks by exploiting loss
  trajectory}. In \bibinfo{booktitle}{\emph{Proceedings of the 2022 ACM SIGSAC
  Conference on Computer and Communications Security}}.
  \bibinfo{pages}{2085--2098}.
\newblock


\bibitem[Lu et~al\mbox{.}(2022)]%
        {lu2022label}
\bibfield{author}{\bibinfo{person}{Zhaobo Lu}, \bibinfo{person}{Hai Liang},
  \bibinfo{person}{Minghao Zhao}, \bibinfo{person}{Qingzhe Lv},
  \bibinfo{person}{Tiancai Liang}, {and} \bibinfo{person}{Yilei Wang}.}
  \bibinfo{year}{2022}\natexlab{}.
\newblock \showarticletitle{Label-only membership inference attacks on machine
  unlearning without dependence of posteriors}.
\newblock \bibinfo{journal}{\emph{International Journal of Intelligent
  Systems}} \bibinfo{volume}{37}, \bibinfo{number}{11} (\bibinfo{year}{2022}),
  \bibinfo{pages}{9424--9441}.
\newblock


\bibitem[Mehnaz et~al\mbox{.}(2022)]%
        {mehnaz2022your}
\bibfield{author}{\bibinfo{person}{Shagufta Mehnaz},
  \bibinfo{person}{Sayanton~V Dibbo}, \bibinfo{person}{Roberta De~Viti},
  \bibinfo{person}{Ehsanul Kabir}, \bibinfo{person}{Bj{\"o}rn~B Brandenburg},
  \bibinfo{person}{Stefan Mangard}, \bibinfo{person}{Ninghui Li},
  \bibinfo{person}{Elisa Bertino}, \bibinfo{person}{Michael Backes},
  \bibinfo{person}{Emiliano De~Cristofaro}, {et~al\mbox{.}}}
  \bibinfo{year}{2022}\natexlab{}.
\newblock \showarticletitle{Are your sensitive attributes private? Novel model
  inversion attribute inference attacks on classification models}. In
  \bibinfo{booktitle}{\emph{31st USENIX Security Symposium (USENIX Security
  22)}}. \bibinfo{pages}{4579--4596}.
\newblock


\bibitem[Nasr et~al\mbox{.}(2018)]%
        {nasr2018machine}
\bibfield{author}{\bibinfo{person}{Milad Nasr}, \bibinfo{person}{Reza Shokri},
  {and} \bibinfo{person}{Amir Houmansadr}.} \bibinfo{year}{2018}\natexlab{}.
\newblock \showarticletitle{Machine learning with membership privacy using
  adversarial regularization}. In \bibinfo{booktitle}{\emph{Proceedings of the
  2018 ACM SIGSAC conference on computer and communications security}}.
  \bibinfo{pages}{634--646}.
\newblock


\bibitem[Rezaei and Liu(2021)]%
        {rezaei2021difficulty}
\bibfield{author}{\bibinfo{person}{Shahbaz Rezaei} {and} \bibinfo{person}{Xin
  Liu}.} \bibinfo{year}{2021}\natexlab{}.
\newblock \showarticletitle{On the difficulty of membership inference attacks}.
  In \bibinfo{booktitle}{\emph{Proceedings of the IEEE/CVF Conference on
  Computer Vision and Pattern Recognition}}. \bibinfo{pages}{7892--7900}.
\newblock


\bibitem[Rezaei and Liu(2022)]%
        {rezaei2022efficient}
\bibfield{author}{\bibinfo{person}{Shahbaz Rezaei} {and} \bibinfo{person}{Xin
  Liu}.} \bibinfo{year}{2022}\natexlab{}.
\newblock \showarticletitle{An efficient subpopulation-based membership
  inference attack}.
\newblock \bibinfo{journal}{\emph{arXiv preprint arXiv:2203.02080}}
  (\bibinfo{year}{2022}).
\newblock


\bibitem[Sagala(2022)]%
        {sagala2022comparative}
\bibfield{author}{\bibinfo{person}{Noviyanti~TM Sagala}.}
  \bibinfo{year}{2022}\natexlab{}.
\newblock \showarticletitle{Comparative Analysis of Grid-based Decision Tree
  and Support Vector Machine for Crime Category Prediction}. In
  \bibinfo{booktitle}{\emph{2021 International Seminar on Machine Learning,
  Optimization, and Data Science (ISMODE)}}. IEEE, \bibinfo{pages}{184--188}.
\newblock


\bibitem[Salem et~al\mbox{.}(2019)]%
        {salem2019ml}
\bibfield{author}{\bibinfo{person}{Ahmed Salem}, \bibinfo{person}{Yang Zhang},
  \bibinfo{person}{Mathias Humbert}, \bibinfo{person}{Pascal Berrang},
  \bibinfo{person}{Mario Fritz}, {and} \bibinfo{person}{Michael Backes}.}
  \bibinfo{year}{2019}\natexlab{}.
\newblock \showarticletitle{ML-Leaks: Model and Data Independent Membership
  Inference Attacks and Defenses on Machine Learning Models}. In
  \bibinfo{booktitle}{\emph{Network and Distributed Systems Security (NDSS)
  Symposium 2019}}.
\newblock


\bibitem[Shokri et~al\mbox{.}(2017)]%
        {shokri2017membership}
\bibfield{author}{\bibinfo{person}{Reza Shokri}, \bibinfo{person}{Marco
  Stronati}, \bibinfo{person}{Congzheng Song}, {and} \bibinfo{person}{Vitaly
  Shmatikov}.} \bibinfo{year}{2017}\natexlab{}.
\newblock \showarticletitle{Membership inference attacks against machine
  learning models}. In \bibinfo{booktitle}{\emph{2017 IEEE symposium on
  security and privacy (SP)}}. IEEE, \bibinfo{pages}{3--18}.
\newblock


\bibitem[Song and Shmatikov(2020)]%
        {song2020overlearning}
\bibfield{author}{\bibinfo{person}{Congzheng Song} {and}
  \bibinfo{person}{Vitaly Shmatikov}.} \bibinfo{year}{2020}\natexlab{}.
\newblock \showarticletitle{Overlearning Reveals Sensitive Attributes}. In
  \bibinfo{booktitle}{\emph{8th International Conference on Learning
  Representations, ICLR 2020}}.
\newblock


\bibitem[Song and Mittal(2021)]%
        {song2021systematic}
\bibfield{author}{\bibinfo{person}{Liwei Song} {and} \bibinfo{person}{Prateek
  Mittal}.} \bibinfo{year}{2021}\natexlab{}.
\newblock \showarticletitle{Systematic Evaluation of Privacy Risks of Machine
  Learning Models.}. In \bibinfo{booktitle}{\emph{USENIX Security Symposium}},
  Vol.~\bibinfo{volume}{1}. \bibinfo{pages}{4}.
\newblock


\bibitem[Vo et~al\mbox{.}(2021)]%
        {vo2021ramboattack}
\bibfield{author}{\bibinfo{person}{Viet~Quoc Vo}, \bibinfo{person}{Ehsan
  Abbasnejad}, {and} \bibinfo{person}{Damith~C Ranasinghe}.}
  \bibinfo{year}{2021}\natexlab{}.
\newblock \showarticletitle{Ramboattack: A robust query efficient deep neural
  network decision exploit}.
\newblock \bibinfo{journal}{\emph{arXiv preprint arXiv:2112.05282}}
  (\bibinfo{year}{2021}).
\newblock


\bibitem[Watson et~al\mbox{.}({[n.\,d.]})]%
        {watsonimportance}
\bibfield{author}{\bibinfo{person}{Lauren Watson}, \bibinfo{person}{Chuan Guo},
  \bibinfo{person}{Graham Cormode}, {and} \bibinfo{person}{Alexandre
  Sablayrolles}.} \bibinfo{year}{[n.\,d.]}\natexlab{}.
\newblock \showarticletitle{On the Importance of Difficulty Calibration in
  Membership Inference Attacks}. In \bibinfo{booktitle}{\emph{International
  Conference on Learning Representations}}.
\newblock


\bibitem[Wexler et~al\mbox{.}(2019)]%
        {wexler2019if}
\bibfield{author}{\bibinfo{person}{James Wexler}, \bibinfo{person}{Mahima
  Pushkarna}, \bibinfo{person}{Tolga Bolukbasi}, \bibinfo{person}{Martin
  Wattenberg}, \bibinfo{person}{Fernanda Vi{\'e}gas}, {and}
  \bibinfo{person}{Jimbo Wilson}.} \bibinfo{year}{2019}\natexlab{}.
\newblock \showarticletitle{The what-if tool: Interactive probing of machine
  learning models}.
\newblock \bibinfo{journal}{\emph{IEEE transactions on visualization and
  computer graphics}} \bibinfo{volume}{26}, \bibinfo{number}{1}
  (\bibinfo{year}{2019}), \bibinfo{pages}{56--65}.
\newblock


\bibitem[Ye et~al\mbox{.}(2022)]%
        {ye2022enhanced}
\bibfield{author}{\bibinfo{person}{Jiayuan Ye}, \bibinfo{person}{Aadyaa Maddi},
  \bibinfo{person}{Sasi~Kumar Murakonda}, \bibinfo{person}{Vincent
  Bindschaedler}, {and} \bibinfo{person}{Reza Shokri}.}
  \bibinfo{year}{2022}\natexlab{}.
\newblock \showarticletitle{Enhanced membership inference attacks against
  machine learning models}. In \bibinfo{booktitle}{\emph{Proceedings of the
  2022 ACM SIGSAC Conference on Computer and Communications Security}}.
  \bibinfo{pages}{3093--3106}.
\newblock


\bibitem[Yeom et~al\mbox{.}(2018)]%
        {yeom2018privacy}
\bibfield{author}{\bibinfo{person}{Samuel Yeom}, \bibinfo{person}{Irene
  Giacomelli}, \bibinfo{person}{Matt Fredrikson}, {and} \bibinfo{person}{Somesh
  Jha}.} \bibinfo{year}{2018}\natexlab{}.
\newblock \showarticletitle{Privacy risk in machine learning: Analyzing the
  connection to overfitting}. In \bibinfo{booktitle}{\emph{2018 IEEE 31st
  computer security foundations symposium (CSF)}}. IEEE,
  \bibinfo{pages}{268--282}.
\newblock


\bibitem[Yu et~al\mbox{.}(2021)]%
        {yu2021reinforcement}
\bibfield{author}{\bibinfo{person}{Chao Yu}, \bibinfo{person}{Jiming Liu},
  \bibinfo{person}{Shamim Nemati}, {and} \bibinfo{person}{Guosheng Yin}.}
  \bibinfo{year}{2021}\natexlab{}.
\newblock \showarticletitle{Reinforcement learning in healthcare: A survey}.
\newblock \bibinfo{journal}{\emph{ACM Computing Surveys (CSUR)}}
  \bibinfo{volume}{55}, \bibinfo{number}{1} (\bibinfo{year}{2021}),
  \bibinfo{pages}{1--36}.
\newblock


\bibitem[Zhang et~al\mbox{.}(2022)]%
        {zhang2022label}
\bibfield{author}{\bibinfo{person}{Guangsheng Zhang}, \bibinfo{person}{Bo Liu},
  \bibinfo{person}{Tianqing Zhu}, \bibinfo{person}{Ming Ding}, {and}
  \bibinfo{person}{Wanlei Zhou}.} \bibinfo{year}{2022}\natexlab{}.
\newblock \showarticletitle{Label-Only Membership Inference Attacks and
  Defenses In Semantic Segmentation Models}.
\newblock \bibinfo{journal}{\emph{IEEE Transactions on Dependable and Secure
  Computing}} (\bibinfo{year}{2022}).
\newblock


\end{thebibliography}

\end{document}